%% file: main.tex
\pdfoutput=1
\documentclass{article} %
\usepackage[T1]{fontenc} %
\usepackage{iclr2027_conference,times}

\input{math_commands.tex}

\input{videoVariables.tex}

\usepackage{graphicx}
\usepackage{subcaption}
\usepackage{float}
\usepackage{hyperref}
\hypersetup{breaklinks=true}
\usepackage{xurl}
\usepackage[capitalise]{cleveref}
\usepackage{url}
\usepackage{videoAbbreviations}	%
\usepackage{enumitem}
\usepackage{booktabs}
\newcommand{\gp}[1]{{\scriptsize\color{gray!70}(#1)}}
\newcommand{\naa}{{\scriptsize\color{gray!70}n/a}}
\usepackage{multirow}
\usepackage{rotating}
\usepackage{videoPlotStyle}	%
\usetikzlibrary{arrows.meta}	%
\usepackage{wrapfig}
\usepackage{placeins}	%
\newcommand{\promptbox}[2]{%
  \par\smallskip\noindent
  \begin{tikzpicture}
    \node[fill=gray!7, rounded corners=2pt, inner xsep=16pt, inner ysep=7pt,
      text width=\linewidth-32pt, font=\small\itshape, align=justify] (p) {#2};
    \node[anchor=north west, inner sep=0pt, text=gray!50,
      font=\fontsize{24}{24}\selectfont] at ([xshift=3pt, yshift=-2pt]p.north west) {\textquotedblleft};
    \node[anchor=south east, inner sep=0pt, text=gray!50,
      font=\fontsize{24}{24}\selectfont] at ([xshift=-3pt, yshift=-8pt]p.south east) {\textquotedblright};
    \node[anchor=south west, inner sep=1pt, text=gray!75,
      font=\scriptsize\scshape] at ([yshift=1pt]p.north west) {#1};
  \end{tikzpicture}\par\smallskip}

\newif\ifcomments
\commentsfalse
\ifcomments
    \newcommand{\jakub}[1]{{\leavevmode\color{blue}JS: #1}}
    \newcommand{\kamil}[1]{{\leavevmode\color{orange}KD: #1}}
    \newcommand{\bartek}[1]{{\leavevmode\color{purple}BT: #1}}
    \newcommand{\jonas}[1]{{\leavevmode\color{teal}JK: #1}}
    \newcommand{\wip}[1]{{\leavevmode\color{gray}WiP: #1}}
    \newcommand{\todo}[1]{{\leavevmode\color{red}todo: #1}}
\else
    \newcommand{\jakub}[1]{}
    \newcommand{\kamil}[1]{}
    \newcommand{\bartek}[1]{}
    \newcommand{\jonas}[1]{}
    \newcommand{\wip}[1]{}
    \newcommand{\todo}[1]{}
\fi

\title{
    Does Physics Live in the Activations? \\ Localizing Physical Quantities in Video Diffusion Models
}
\iclrfinalcopy

\author{
Jonas Kneifl$^{1}$ \quad
Jakub Skalski$^{1}$ \quad
Bart{\l}omiej Twardowski$^{1,2}$ \quad
Kamil Deja$^{1,3}$ \\[2pt]
$^{1}$IDEAS Research Institute, Warsaw, Poland \\
$^{2}$Computer Vision Center, Universitat Aut\`onoma de Barcelona, Spain \\
$^{3}$Warsaw University of Technology, Poland \\[2pt]
\texttt{\{jonas.kneifl, jakub.skalski,} \\
\texttt{bartlomiej.twardowski, kamil.deja\}@ideas.edu.pl}
}

\begin{document}

\maketitle

\begin{abstract}

Video generation models produce strikingly realistic sequences and are increasingly proposed as world models, yet recent benchmarks reveal pronounced deficits in their physical reasoning.
This raises the question of whether these models internalize physical principles or merely reproduce familiar motion patterns.
We address this by probing internal representations of video \glspl{dit} for simulator-derived ground-truth physical quantities spanning kinematic motion and rigid-body dynamics under gravity and contact.
We find that these quantities are linearly decodable with high accuracy early in the denoising process, substantially outperforming a baseline decoded directly from the model's own noised latents---indicating that the relevant physical information is actively constructed during denoising, rather than already present in the input.

Additionally, we show that activations at on-object tokens carry the relevant physical information and that quantities defined over multiple frames are readable from single latent frames. 
Hence, information is sharply localized within the token sequence and is computed globally but stored locally. 
The probes further show partial extrapolation, transferring to scene variations and object configurations outside their training regime, so what they read is not simply a correlate of the scenes they were fit on.
When fitted directly in the full-resolution activation space, the probing directions can serve as steering vectors to change the model's output.
\end{abstract}

\input{sections/introduction.tex}

\input{sections/related_work.tex}

\input{sections/setup.tex}

\input{sections/probing.tex}

\input{sections/discussion.tex}

\input{sections/conclusion.tex}

\bibliography{iclr2027_conference}
\bibliographystyle{iclr2027_conference}

\appendix
\raggedbottom
\counterwithin{figure}{section}
\counterwithin{table}{section}
\input{sections/appendix.tex}

\end{document}

%% file: math_commands.tex
\usepackage{amsmath,amsfonts,bm}

\def\eqref#1{equation~\ref{#1}}

\def\1{\bm{1}}

\DeclareMathAlphabet{\mathsfit}{\encodingdefault}{\sfdefault}{m}{sl}
\SetMathAlphabet{\mathsfit}{bold}{\encodingdefault}{\sfdefault}{bx}{n}

\newcommand{\R}{\mathbb{R}}

%% file: sections/introduction.tex
\label{sec:intro}

Video generation models~\citep{yangCogVideoXTexttoVideoDiffusion2024,wanWanOpenAdvanced2025, zhengOpenSora20Training2026} can produce remarkably realistic motion, and show impressive results in vision understanding tasks~\citep{wiedemerVideoModelsAre2025,wangDemystifyingVideoReasoning2026}. This makes them natural candidates for general-purpose world models~\citep{brooks2024video,yang2024video}.
Yet recent evaluations show that this visual realism does not imply reliable physical behavior: generated videos can look plausible while violating basic physical constraints~\citep{mPhyEduVideoBenchmarkEvaluating2026,guPhyWorldBenchComprehensiveEvaluation2025,guoT2VPhysBenchFirstPrinciplesBenchmark2025}.
This raises an important question that output-level evaluations cannot answer: when a video model produces physically consistent motion, does it do so because it represents the underlying physical quantities, or does it simply reproduce patterns that are common in its training data?

\begin{figure}[t]
  \centering
  \input{figures/appetizer/appetizer.tex}
  \caption{What a frozen video \gls{dit} knows about physics. 
  For each scenario (left), the physical signal concentrates on the object's tokens, which the probe pools as a disk-shaped \acrshort{roi} (\emph{localization};
  schematic heat, not a measured map). 
  A linear probe recovers speed, total momentum~$\mom$ and gravity from the
  residual stream along the denoising trajectory (\emph{probing}), and its direction holds up downstream (right):
  it still reads the quantity once the ball is replaced by a can, while a full-resolution probe injected back flips the rolling direction.
  }
  \label{fig:overview}
\end{figure}
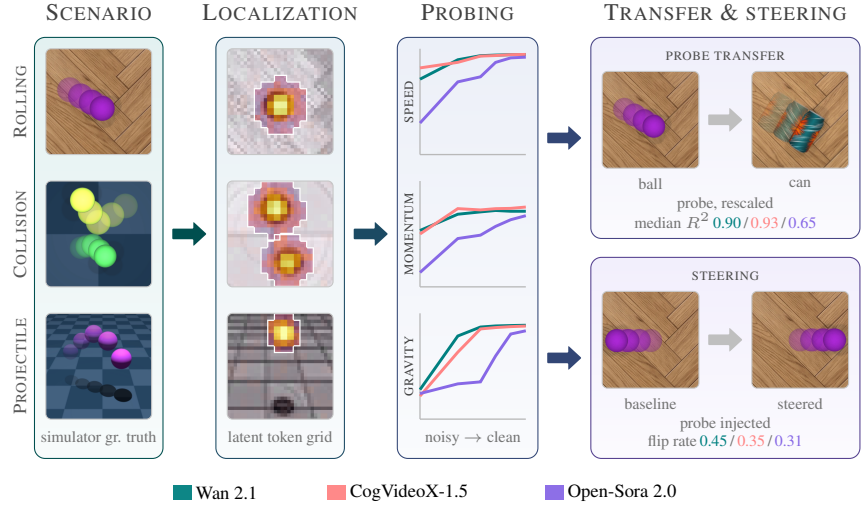

We study this question from the perspective of the model's internal representations as illustrated in~\cref{fig:overview}. Rather than asking whether a generated video looks physically plausible, we ask whether the model's activations contain explicit information about the physical state of the scene. In particular, following the Linear Representation Hypothesis, we investigate whether quantities such as speed, direction, momentum, and gravitational acceleration are represented in a form that can be linearly decoded from the hidden states of video \glspl{dit} such as Wan~2.1~\citep{wanWanOpenAdvanced2025}, CogVideoX-1.5~\citep{yangCogVideoXTexttoVideoDiffusion2024}, and Open-Sora~2.0~\citep{zhengOpenSora20Training2026}.

To make this question measurable, we render controlled scenarios with a physics simulator to gather exact ground-truth values for every physical quantity of interest. We encode these videos, forward-diffuse them to a range of noise levels, and record the activations of a single frozen \gls{dit} forward pass. Ridge probes then predict the simulator-derived quantities across layers, noise levels, and token subsets. 
We compare those results against probes applied directly to the model's noised latent and image conditioning, allowing us to distinguish information already present in the input from information constructed inside the transformer.

In all three models, physical quantities are strongly linearly decodable from the activations and well beyond what the input carries: quantities such as direction become readable long before the input latent reveals it, and gravity, which the latent does not reveal even when clean, is decoded precisely from the activations.
Hence, physical information is not simply read from the models' input, but is constructed inside the transformer during denoising.
The information is highly localized --- a \gls{roi} covering 1--2\% of the tokens around the object beats a global readout over the whole token sequence --- and temporally structured, with each token-frame favoring the physical state of its own phase.
Speed and gravity are readable from a single token-frame, even though they require information gathered across time. 
Thus, the transformer aggregates information across the sequence and stores it locally around the object.
The probes transfer to variations of background, object shape and count, as well as perspective, suggesting that the decoded signal captures physical structure rather than only scene-specific appearance. 
Finally, we ask whether model-internal representations are causally active in generation.
For the rolling direction, full-resolution probes, when injected back, can change the generated motion.

Our contributions can be summarized as follows:
\begin{itemize}
    \item We establish that physical quantities are linearly decodable from video \gls{dit} activations.
    \item We find that those properties are computed across the generated video, but stored locally around the relevant objects.
    \item We show that representations encoding physical quantities transfer across scene variations and are causally active in generation.
\end{itemize}

%% file: figures/appetizer/appetizer.tex
\input{figures/appetizer/probing.tex}%
\newif\ifappLegendRight%
\appLegendRightfalse%
\newif\ifappMergedArrows%
\appMergedArrowsfalse%
\def\appHeatVariant{_roi_dim}%
\def\appTile{1.4cm}%
\def\appRowSep{1.75cm}%
\def\appColScene{0cm}%
\def\appColHeat{2.4cm}%
\def\appColProbe{4.95cm}%
\def\appColProbePanel{4.88cm}%
\def\appPanelHalf{0.85cm}%
\def\appProbeHalf{0.93cm}%
\ifappMergedArrows%
  \def\appColDown{8.21cm}%
  \def\appDownHalf{1.7cm}%
  \def\appDownTopBot{-1.25cm}%
  \def\appDownBotTop{-2.25cm}%
  \def\appDownTile{1.28cm}%
  \def\appDownFrameDx{0.99cm}%
  \def\appDownSubOff{0.21cm}%
  \def\appDownImgOff{0.98cm}%
  \def\appDownTxtOffA{1.68cm}%
  \def\appDownTxtOffB{1.89cm}%
\else%
  \def\appColDown{8.29cm}%
  \def\appDownHalf{1.78cm}%
  \def\appDownTopBot{-1.81cm}%
  \def\appDownBotTop{-2.06cm}%
  \def\appDownTile{1.28cm}%
  \def\appDownFrameDx{0.99cm}%
  \def\appDownSubOff{0.24cm}%
  \def\appDownImgOff{1.09cm}%
  \def\appDownTxtOffA{2.21cm}%
  \def\appDownTxtOffB{2.42cm}%
\fi%
\def\appPanelTop{0.85cm}%
\def\appPanelBot{-4.72cm}%
\def\appMidY{-1.75cm}%
\def\appHeaderY{0.95cm}%
\pgfplotsset{appProbeAxis/.append style={width=\appTile, height=\appTile}}%
\colorlet{appTintScene}{modelWan}%
\colorlet{appTintLoc}{modelWan!68!modelSora}%
\colorlet{appTintProbe}{modelWan!40!modelSora}%
\colorlet{appTintTrans}{modelWan!15!modelSora}%
\colorlet{appTintSteer}{modelSora}%
\newcommand{\appPanelYY}[5]{%
  \shadedraw[rounded corners=4pt, draw=#5!62!black, line width=0.4pt,%
             top color=#5!11, bottom color=#5!3]%
    ($(#1,#3)-(#2,0)$) rectangle ($(#1,#4)+(#2,0)$);}%
\newcommand{\appPanelArrowYY}[6]{%
  \shadedraw[rounded corners=4pt, draw=#5!62!black, line width=0.4pt,%
             top color=#5!11, bottom color=#5!3]%
    ($(#1,#3)-(#2,0)$) -- ($(#1,#3)+(#2,0)$)%
    [rounded corners=0.4pt]%
    -- ($(#1,\appMidY)+(#2,2.2pt)$)%
    -- ($(#6,\appMidY)+(-6.5pt,2.2pt)$)%
    -- ($(#6,\appMidY)+(-6.5pt,6pt)$)%
    -- ($(#6,\appMidY)$)%
    -- ($(#6,\appMidY)+(-6.5pt,-6pt)$)%
    -- ($(#6,\appMidY)+(-6.5pt,-2.2pt)$)%
    -- ($(#1,\appMidY)+(#2,-2.2pt)$)%
    [rounded corners=4pt]%
    -- ($(#1,#4)+(#2,0)$) -- ($(#1,#4)-(#2,0)$) -- cycle;}%
\newcommand{\appPanelForkYY}[8]{%
  \shadedraw[rounded corners=4pt, draw=#5!62!black, line width=0.4pt,%
             top color=#5!11, bottom color=#5!3]%
    ($(#1,#3)-(#2,0)$) -- ($(#1,#3)+(#2,0)$)%
    [rounded corners=0.4pt]%
    -- ($(#1,\appMidY)+(#2,2.2pt)$)%
    -- ($(#6,\appMidY)+(-2.2pt,2.2pt)$)%
    -- ($(#6,#7)+(-2.2pt,-6.5pt)$)%
    -- ($(#6,#7)+(-6pt,-6.5pt)$)%
    -- ($(#6,#7)$)%
    -- ($(#6,#7)+(6pt,-6.5pt)$)%
    -- ($(#6,#7)+(2.2pt,-6.5pt)$)%
    -- ($(#6,#8)+(2.2pt,6.5pt)$)%
    -- ($(#6,#8)+(6pt,6.5pt)$)%
    -- ($(#6,#8)$)%
    -- ($(#6,#8)+(-6pt,6.5pt)$)%
    -- ($(#6,#8)+(-2.2pt,6.5pt)$)%
    -- ($(#6,\appMidY)+(-2.2pt,-2.2pt)$)%
    -- ($(#1,\appMidY)+(#2,-2.2pt)$)%
    [rounded corners=4pt]%
    -- ($(#1,#4)+(#2,0)$) -- ($(#1,#4)-(#2,0)$) -- cycle;}%
\newcommand{\appPanel}[3]{\appPanelYY{#1}{#2}{\appPanelTop}{\appPanelBot}{#3}}%
\newcommand{\appHeader}[2]{%
  \node[anchor=south, font=\footnotesize\scshape, text=black!75] at (#1,\appHeaderY) {#2};}%
\newcommand{\appTileImgWH}[6]{%
  \node[inner sep=0pt, minimum width=#4, minimum height=#5,%
        rounded corners=2pt, draw=gray!45, line width=0.4pt,%
        path picture={\node at (path picture bounding box.center)%
          {\includegraphics[width=#4]{#6}};}] (#1) at (#2,#3) {};}%
\newcommand{\appTileImg}[4]{\appTileImgWH{#1}{#2}{#3}{\appTile}{\appTile}{#4}}%
\newcommand{\appArrowXX}[4]{%
  \fill[#4, rounded corners=0.4pt]%
    ($(#1,#3)+(0,2pt)$) -- ($(#2,#3)+(-5pt,2pt)$) -- ($(#2,#3)+(-5pt,4.6pt)$) --%
    ($(#2,#3)$) -- ($(#2,#3)+(-5pt,-4.6pt)$) -- ($(#2,#3)+(-5pt,-2pt)$) --%
    ($(#1,#3)-(0,2pt)$) -- cycle;}%
\newcommand{\appForkXX}[6]{%
  \fill[#6, rounded corners=0.4pt]%
    ($(#1,#5)+(0,2pt)$) -- ($(#2,#5)+(-2pt,2pt)$) -- ($(#2,#3)+(-2pt,-5pt)$) --%
    ($(#2,#3)+(-4.6pt,-5pt)$) -- ($(#2,#3)$) -- ($(#2,#3)+(4.6pt,-5pt)$) --%
    ($(#2,#3)+(2pt,-5pt)$) -- ($(#2,#4)+(2pt,5pt)$) -- ($(#2,#4)+(4.6pt,5pt)$) --%
    ($(#2,#4)$) -- ($(#2,#4)+(-4.6pt,5pt)$) -- ($(#2,#4)+(-2pt,5pt)$) --%
    ($(#2,#5)+(-2pt,-2pt)$) -- ($(#1,#5)-(0,2pt)$) -- cycle;}%
\newcommand{\appProbeChip}[2]{%
  \begin{scope}[shift={(#1,#2)}]%
    \draw[fill=white, draw=gray!50, rounded corners=1.2pt, line width=0.3pt]%
      (-7.5pt,-3.2pt) rectangle (7.5pt,3.2pt);%
    \foreach \i/\t in {0/75, 1/25, 2/95, 3/15, 4/55} {%
      \fill[modelWan!\t] ($(-6.5pt,-2pt)+(\i*2.6pt,0)$) rectangle +(2pt,4pt);}%
  \end{scope}}%
\newcommand{\appTokenChip}[2]{%
  \begin{scope}[shift={(#1,#2)}]%
    \draw[fill=white, draw=gray!50, rounded corners=1.2pt, line width=0.3pt]%
      (-7.5pt,-5.6pt) rectangle (7.5pt,5.6pt);%
    \foreach \i/\r in {0/85, 1/35, 2/95, 3/20, 4/60} {%
      \foreach \j/\d in {0/0, 1/-25, 2/10} {%
        \pgfmathtruncatemacro{\cellt}{max(12,min(95,\r+\d+12*\j))}%
        \fill[modelWan!\cellt] ($(-6.5pt,-4.4pt)+(\i*2.6pt,\j*3pt)$) rectangle +(2pt,2.2pt);}}%
  \end{scope}}%
\newcommand{\appRow}[5]{%
  \pgfmathsetlengthmacro{\appY}{-#1*\appRowSep}%
  \appTileImg{scene#1}{\appColScene}{\appY}{figures/appetizer/#2.png}%
  \appTileImg{heat#1}{\appColHeat}{\appY}{figures/appetizer/#2_heatmap\appHeatVariant.png}%
  \node[anchor=south, font=\scriptsize\scshape, text=black!65, rotate=90, inner sep=7pt]%
    at (scene#1.west) {#5};%
  \appProbePanel{#3}{#4}{at={(\appColProbe,\appY)}, anchor=center}}%
\input{figures/appetizer/transfer_shape.tex}
\input{figures/appetizer/steering_flip.tex}%
\begin{tikzpicture}[font=\scriptsize]
  \def\appArrowGap{3.5pt}
  \pgfmathsetlengthmacro{\appHeatL}{\appColHeat-\appPanelHalf-\appArrowGap}
  \pgfmathsetlengthmacro{\appProbeL}{\appColProbePanel-\appProbeHalf-\appArrowGap}
  \pgfmathsetlengthmacro{\appForkTop}{\appDownTopBot-\appArrowGap}
  \pgfmathsetlengthmacro{\appForkBot}{\appDownBotTop+\appArrowGap}

  \ifappMergedArrows
    \appPanelArrowYY{\appColScene}{\appPanelHalf}{\appPanelTop}{\appPanelBot}{appTintScene}{\appHeatL}
    \appPanelArrowYY{\appColHeat}{\appPanelHalf}{\appPanelTop}{\appPanelBot}{appTintLoc}{\appProbeL}
    \appPanelForkYY{\appColProbePanel}{\appProbeHalf}{\appPanelTop}{\appPanelBot}{appTintProbe}{\appColDown}{\appForkTop}{\appForkBot}
  \else
    \appPanel{\appColScene}{\appPanelHalf}{appTintScene}
    \appPanel{\appColHeat}{\appPanelHalf}{appTintLoc}
    \appPanel{\appColProbePanel}{\appProbeHalf}{appTintProbe}
    \pgfmathsetlengthmacro{\appSceneR}{\appColScene+\appPanelHalf+\appArrowGap}
    \pgfmathsetlengthmacro{\appHeatR}{\appColHeat+\appPanelHalf+\appArrowGap}
    \pgfmathsetlengthmacro{\appProbeR}{\appColProbePanel+\appProbeHalf+\appArrowGap}
    \appArrowXX{\appSceneR}{\appHeatL}{\appMidY}{appTintScene!62!black}
    \appArrowXX{\appHeatR}{\appProbeL}{\appMidY}{appTintLoc!62!black}
    \pgfmathsetlengthmacro{\appDownTopMid}{(\appPanelTop+\appDownTopBot)/2}
    \pgfmathsetlengthmacro{\appDownBotMid}{(\appDownBotTop+\appPanelBot)/2}
    \pgfmathsetlengthmacro{\appDownL}{\appColDown-\appDownHalf-\appArrowGap}
    \appArrowXX{\appProbeR}{\appDownL}{\appDownTopMid}{appTintProbe!62!black}
    \appArrowXX{\appProbeR}{\appDownL}{\appDownBotMid}{appTintProbe!62!black}
  \fi
  \appPanelYY{\appColDown}{\appDownHalf}{\appPanelTop}{\appDownTopBot}{appTintTrans}
  \appPanelYY{\appColDown}{\appDownHalf}{\appDownBotTop}{\appPanelBot}{appTintSteer}

  \appRow{0}{rolling}{roll_speed}{speed}{Rolling}
  \appRow{1}{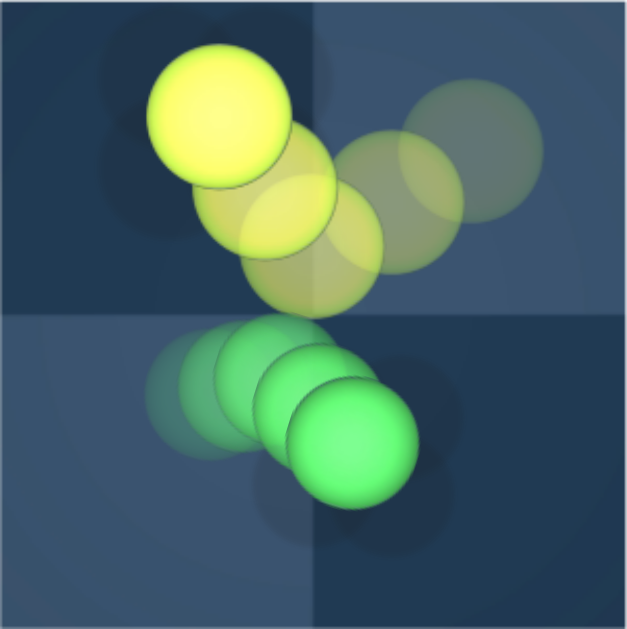}{coll_total_momentum_xy}{momentum}{Collision}
  \appRow{2}{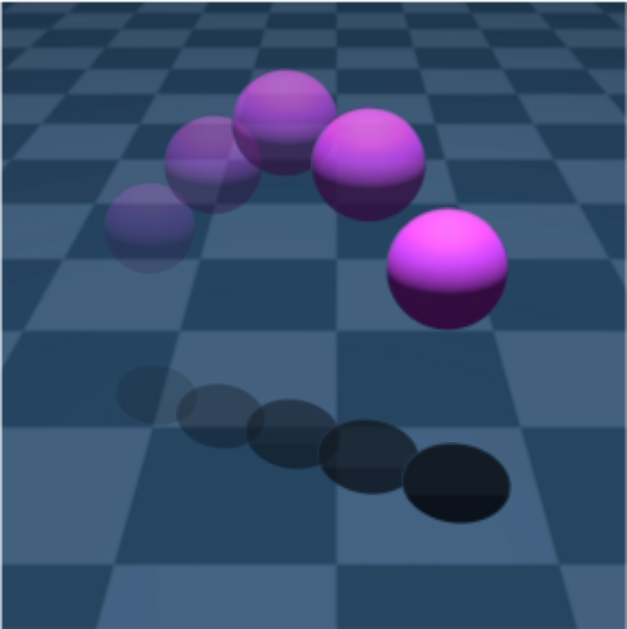}{proj_gravity}{gravity}{Projectile}

  \node[anchor=center, font=\tiny, text=black!55]
    at ($(\appColProbe,\appPanelBot)+(0,0.26cm)$) {noisy $\rightarrow$ clean};
  \node[anchor=center, font=\tiny, text=black!55]
    at ($(\appColScene,\appPanelBot)+(0,0.26cm)$) {simulator gr. truth};
  \node[anchor=center, font=\tiny, text=black!55]
    at ($(\appColHeat,\appPanelBot)+(0,0.26cm)$) {latent token grid};

  \pgfmathsetlengthmacro{\appTrSub}{\appPanelTop-\appDownSubOff}
  \pgfmathsetlengthmacro{\appTrImg}{\appPanelTop-\appDownImgOff}
  \pgfmathsetlengthmacro{\appTrTxtA}{\appPanelTop-\appDownTxtOffA}
  \pgfmathsetlengthmacro{\appTrTxtB}{\appPanelTop-\appDownTxtOffB}
  \def\appDownLabGap{0.2cm}
  \pgfmathsetlengthmacro{\appTrLab}{\appTrImg-\appDownTile/2-\appDownLabGap}
  \pgfmathsetlengthmacro{\appStSub}{\appDownBotTop-\appDownSubOff}
  \pgfmathsetlengthmacro{\appStImg}{\appDownBotTop-\appDownImgOff}
  \pgfmathsetlengthmacro{\appStTxtA}{\appDownBotTop-\appDownTxtOffA}
  \pgfmathsetlengthmacro{\appStTxtB}{\appDownBotTop-\appDownTxtOffB}
  \pgfmathsetlengthmacro{\appStLab}{\appStImg-\appDownTile/2-\appDownLabGap}
  \pgfmathsetlengthmacro{\appDownArrowL}{\appColDown-\appDownFrameDx+\appDownTile/2+\appArrowGap}
  \pgfmathsetlengthmacro{\appDownArrowR}{\appColDown+\appDownFrameDx-\appDownTile/2-\appArrowGap}

  \node[anchor=center, font=\tiny\scshape, text=black!70] at (\appColDown,\appTrSub)
    {probe transfer};
  {\def\appTile{\appDownTile}
   \appTileImg{tsrc}{\appColDown-\appDownFrameDx}{\appTrImg}{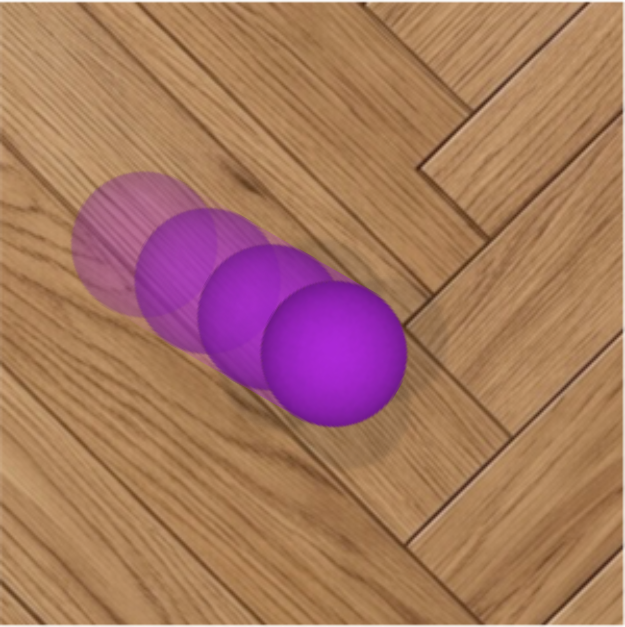}
   \appTileImg{ttgt}{\appColDown+\appDownFrameDx}{\appTrImg}{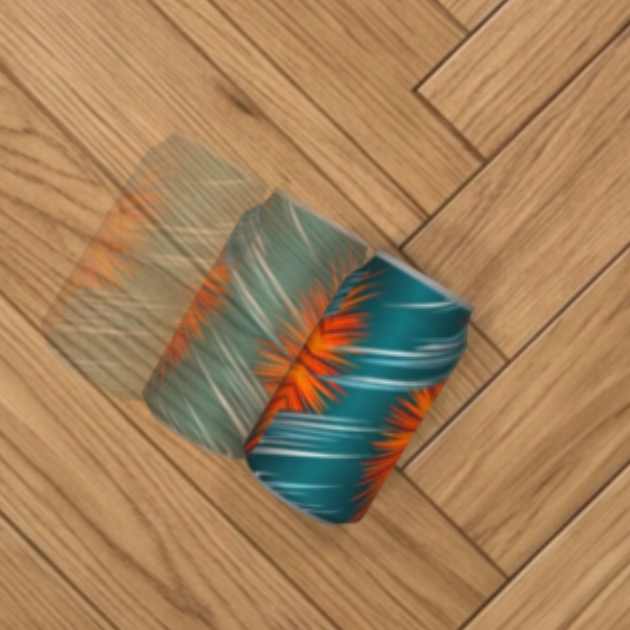}}
  \appArrowXX{\appDownArrowL}{\appDownArrowR}{\appTrImg}{gray!50}
  \node[anchor=center, font=\tiny, text=black!55] at ($(\appColDown,\appTrLab)-(\appDownFrameDx,0)$) {ball};
  \node[anchor=center, font=\tiny, text=black!55] at ($(\appColDown,\appTrLab)+(\appDownFrameDx,0)$) {can};
  \node[anchor=center, font=\tiny, text=black!55] at (\appColDown,\appTrTxtA)
    {probe, rescaled};
  \node[anchor=center, font=\tiny, text=black!55] at (\appColDown,\appTrTxtB)
    {median $\rsq$\,\textcolor{modelWan}{\appShapeWan}\,/\,\textcolor{modelCog}{\appShapeCog}\,/\,\textcolor{modelSora}{\appShapeSora}};

  \node[anchor=center, font=\tiny\scshape, text=black!70] at (\appColDown,\appStSub)
    {steering};
  {\def\appTile{\appDownTile}
   \appTileImg{sneg}{\appColDown-\appDownFrameDx}{\appStImg}{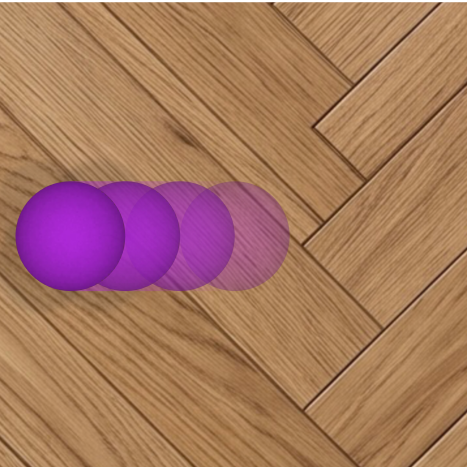}
   \appTileImg{spos}{\appColDown+\appDownFrameDx}{\appStImg}{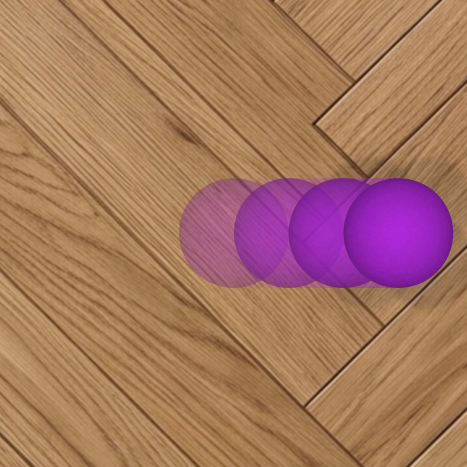}}
  \appArrowXX{\appDownArrowL}{\appDownArrowR}{\appStImg}{gray!50}
  \node[anchor=center, font=\tiny, text=black!55] at ($(\appColDown,\appStLab)-(\appDownFrameDx,0)$) {baseline};
  \node[anchor=center, font=\tiny, text=black!55] at ($(\appColDown,\appStLab)+(\appDownFrameDx,0)$) {steered};
  \node[anchor=center, font=\tiny, text=black!55] at (\appColDown,\appStTxtA)
    {probe injected};
  \node[anchor=center, font=\tiny, text=black!55] at (\appColDown,\appStTxtB)
    {\vphantom{$\rsq$}flip rate\,\textcolor{modelWan}{\appFlipWan}\,/\,\textcolor{modelCog}{0.35}\,/\,\textcolor{modelSora}{0.31}};

  \appHeader{\appColScene}{Scenario}
  \appHeader{\appColHeat}{Localization}
  \appHeader{\appColProbePanel}{Probing}
  \appHeader{\appColDown}{Transfer \& steering}

  \ifappLegendRight
    \begin{scope}[shift={($(\appColDown,\appPanelBot)+(\appDownHalf,0)+(0.54cm,0)$)}]
      \appProbeLegendV
    \end{scope}
  \else
    \begin{scope}[shift={($(\appColScene,\appPanelBot)+(0.98cm,-0.55cm)$)}]
      \appProbeLegend
    \end{scope}
  \fi
\end{tikzpicture}

%% file: sections/related_work.tex
\section{Related Work}
\label{sec:related}

\paragraph{Physics benchmarks}
A growing body of work quantifies the physical correctness of video models.
Intuitive-physics benchmarks test plausibility judgments in vision models~\citep{riochetIntPhys2019Benchmark2022,bordesIntPhys2Benchmarking2025}.
For generative models, one line rates \gls{t2v} outputs through perceptual metrics~\citep{mengWorldSimulatorCrafting2024,mPhyEduVideoBenchmarkEvaluating2026}, binary judgments~\citep{guPhyWorldBenchComprehensiveEvaluation2025} or human ratings of physical adherence~\citep{guoT2VPhysBenchFirstPrinciplesBenchmark2025}, while another matches generations against tracked ground truth, in simulation~\citep{jainRIGIDBENCHEVALUATINGRIGIDBODY2026} and in real footage~\citep{tragoudarasEvaluatingNewtonianMechanics2025,motamedGenerativeVideoModels2025},
or trains \glspl{dit} on synthetic 2D mechanics to test sequence completion~\citep{kangHowFarVideo2025}.
Others compare outcome distributions of stochastic events against analytical references~\citep{sadeghiCaliBenchAreStochastic2026}, or assess symbolic reasoning through overlays of physical quantities~\citep{yaoApple$pi$BenchmarkingThinking2026}.
The picture is consistent: generations look plausible but fail under closer inspection~\citep{guoT2VPhysBenchFirstPrinciplesBenchmark2025,yaoApple$pi$BenchmarkingThinking2026,guPhyWorldBenchComprehensiveEvaluation2025,tragoudarasEvaluatingNewtonianMechanics2025,motamedGenerativeVideoModels2025,sadeghiCaliBenchAreStochastic2026}.
While some benchmarks find proprietary models outperforming open-source ones~\citep{yaoApple$pi$BenchmarkingThinking2026}, others report mixed rankings~\citep{guPhyWorldBenchComprehensiveEvaluation2025,jainRIGIDBENCHEVALUATINGRIGIDBODY2026,
guoT2VPhysBenchFirstPrinciplesBenchmark2025}.

\paragraph{Physics probing in video models}
While the previously mentioned works evaluate models solely through their outputs, none examine internal representations.
A parallel line of work addresses this gap through the lens of mechanistic interpretability, probing video models' internal representations for physical quantities or plausibility.
\citet{josephInterpretingPhysicsVideo2026} find that quantities such as speed and acceleration become linearly accessible at roughly one-third network depth in \gls{vjepa} and \gls{vmae}.
\citet{balloutPixelsPrinciplesProbing2025} show that the vision encoder in \glspl{mllm} forms representations from which physical plausibility is decodable, but this capability does not transfer through vision–language projection.
\citet{punzoVideoFoundationModels2026} compare plausibility representations across video foundation models, spanning encoder architectures and \glspl{dit}, finding \gls{vjepa} to perform best.
In contrast, \citet{esmatiInvisibleHandPhysics2026a} report that \glspl{dit} outperform encoder-based architectures on a plausibility readout.
Additionally, they show that causal influence and linear decodability are dissociated across network depth. 
\citet{liWhyVideoDiffusion2026} instead ask why physics \emph{fails}, tracing \gls{t2v} failures to attention heads whose \gls{rope} decay locks motion in early.
\citet{tangSeekingPhysicsDiffusion2026} exploit probes during video generation in \glspl{dit} as a ranking signal to prune implausible video candidates early in the diffusion process. 

The majority of prior work asks \textit{is it plausible?} We instead probe video \glspl{dit} directly against ground-truth continuous physical quantities, reframing the question as \textit{does it know?}
The works most aligned with this goal either operate on video encoders rather than generative models~\citep{josephInterpretingPhysicsVideo2026}, 
or primarily assess physical plausibility, with only a limited side experiment probing quantitative targets in smaller models and simple 2D animations~\citep{esmatiInvisibleHandPhysics2026a}.

\paragraph{Steering in video models}
\label{sec:related:steering}
Activation steering --- adding a direction to a model's internal state at inference time to change its output --- is well established for \glspl{llm} and is beginning to be adapted to video \glspl{dit}: mostly for safety, with a single supervised direction~\citep{kunduPullingREINSTrainingFree2026} or closed-loop control in a contrastive subspace~\citep{hongActivationSteeringVideo2026}, but also for motion control through test-time latent optimization~\citep{xiaoMOFTVideoDiffusion2024}, for generation quality by renormalizing rare large-magnitude activations~\citep{chengSteeringVideoDiffusion2026}, and for multi-event generation by fusing prompts along the denoising trajectory~\citep{liaoTunerDiTTrainingfreeProgressive2026}.

%% file: sections/setup.tex
\section{Setup}
\label{sec:setup}

\subsection{Synthetic Data Generation}
\label{sec:setup:data}
Real-world video is difficult to annotate with exact physical labels. We therefore use the Genesis physics simulator~\citep{teamRoleSimulationScalable2026} to generate ground-truth videos with exact analytic or simulator-derived annotations for every physical observable of interest.

We consider three scenarios covering complementary classes of observables: \textbf{(i) Rolling without slipping} for kinematics (speed, direction); \textbf{(ii) Elastic collision} of multiple balls for interaction and a conserved quantity, total momentum, and pre-/post-collision phases providing time-varying targets; and \textbf{(iii) Projectile motion} for gravitational acceleration and launch angle, neither of which is detectable from any single frame.
We render 200 videos per scenario, varying appearance and initial conditions. Scenario details, generation parameters and the released dataset are described in \cref{sec:simulation:scenarios}.

\subsection{Models and Activation Capture}
\label{sec:setup:models}

We use the largest open-weight \gls{i2v} variants of three \gls{sota} architectures: Wan~2.1~(14B)~\citep{wanWanOpenAdvanced2025}, whose plain architecture, unlike its \gls{moe}-based successor, avoids expert-routing confounds; CogVideoX-1.5~(5B)~\citep{yangCogVideoXTexttoVideoDiffusion2024}; and Open-Sora~2.0~(11B)~\citep{zhengOpenSora20Training2026} in its Hunyuan-\gls{vae} release~\citep{kongHunyuanVideoSystematicFramework2025}, which keeps compression rate and \gls{vae} architecture comparable across models, see~\cref{sec:appendix:models}.

All three models are \glspl{dit} that denoise compressed video latents.
A \gls{vae} maps a clip $\video \in \R^{(\nframes+1) \timesSmall \height \timesSmall \width \timesSmall \channels}$, with $\nframes+1$ frames of spatial resolution $\height \times \width$ and $\channels$ channels, to a latent representation $\latent \in \R^{(\nframesLat+1) \timesSmall \heightLat \timesSmall \widthLat \timesSmall \channelsLat}$. 
The first frame is encoded on its own, while each subsequent group of four frames is encoded jointly, resulting in roughly $4\times$ temporal compression with $\nframes = 4\nframesLat$. 
Independently, the \gls{vae} provides $8\times$ spatial compression.
The latent is split into $p_t \timesSmall p_h \timesSmall p_w$ spatio-temporal patches, flattened row-major into $\ntokens$ tokens, and linearly embedded in the residual stream $\resid_{\layer} \in \R^{\ntokens \timesSmall \dmodel}$ processed by $\nlayers$ \gls{dit} blocks $\layer \in \{0,\dots,\nlayers-1\}$.
Because attention spans the full sequence, every token is contextualized by the entire clip.
Each token index~$\tokenIndex$ has a fixed latent position~$(\frame, h, w)$, a mapping our localization experiments read in reverse.
We call the tokens with frame index~$\frame$ a \emph{token-frame}, arranged as a $\heightLat/p_h \timesSmall \widthLat/p_w$ \emph{token grid}.

\paragraph{Activation Capture}
The models are trained to make single-step predictions at arbitrary noise levels.
Rather than sample from pure noise, we encode a labeled ground-truth video~$\video_{\gt}$ as $\clean=\encoder(\video_{\gt})$, diffuse it to a chosen noise level~$\dstep$ to obtain~$\noisy$, and run the denoising pass from there.
We record each layer's activations $\actAt{\layer}{\dstep} \in \R^{\ntokens \timesSmall \dmodel}$.
This keeps inference within the training distribution while maintaining exact physical labels.
Videos use each model's native resolution and frame count, and the resulting activation--label pairs train the probes.
During capture, we condition on the first ground-truth frame and use an empty prompt to avoid a mismatch between ground-truth video and text conditioning.

%% file: sections/probing.tex
\section{Probing Physical Observables}
\label{sec:probing}

\subsection{Method}
\label{sec:probing:method}
\paragraph{Estimator}
Probes are ridge regressions on standardized features, with both the standardization statistics and the penalty~$\ridge$ fit on training folds only.
We report~$\rsq$ under $k$-fold cross-validation ($k=5$) with folds assigned at the video level, so no frame of a held-out video is seen during fitting.
For vector-valued targets (direction, total momentum) we average the per-component~$\rsq$ uniformly.
Probes are trained on the residual stream~$\resid_\layer$.
Noise levels are reported as the \gls{snr} of the noised latent in\,dB, which, unlike a denoising step or diffusion timestep, is comparable across models with different noise schedules.
A permutation null, refitting every probe on targets shuffled across videos, leaves~$\rsq$ at chance throughout (mean~$-0.056$; \cref{sec:appendix:protocol}), validating the probing protocol.

\paragraph{Pooling and localization}
Time-varying targets are probed from spatially pooled activations, $\meanOf{\act}_{\text{sp}}\in\R^{(\nframesLat+1)\times\dmodel}$, retaining the temporal latent-frame axis so that each token-frame serves as a regression sample; targets constant over the clip are instead probed from the spatio-temporal mean, ${\meanOf{\act}}_{\text{st}}\in\R^{\dmodel}$ (\cref{sec:appendix:protocol}).
Orthogonally, we vary which tokens are pooled: either all tokens (the \emph{global} readout) or only a disk-shaped \gls{roi} centered on the object's simulator-derived position projected onto the latent grid.
The disk's radius is set relative to the projected object radius in each latent frame (default factor 1.5).
Both token sets are mean-pooled, so the two readouts differ only in which tokens enter the mean, not in feature dimension or estimator.
Unless stated otherwise we report the \gls{roi} readout.

\paragraph{Input baseline}
The model's input already carries physical signal, from the noised latent~$\noisy$ (\cref{sec:setup:models}) and from the clean conditioning image.
As a reference we therefore probe the input itself: the baseline~$\baseline$ concatenates~$\noisy$ with the conditioning latent and is pooled exactly like the activations (\cref{sec:appendix:protocol}).

\subsection{Physical Quantities Are Linearly Decodable}
\label{sec:probing:results}
\begin{figure}[t]
  \centering
  \input{figures/probing/decodability_lines_sweep.tex}
  \caption{Probe decodability ($\rsq$, $n=200$ videos, $5$-fold video-level cross-validation) from \gls{dit} activations at the \gls{roi}, across noise levels (\gls{snr} of the noised latent). Solid colored lines are the mean
  over transformer blocks, bands their min--max spread (10 blocks uniformly across network depth). 
  Gray: the latent references, \gls{roi} masked the same way --- solid $\baseline$ ($\noisy$ with the clean conditioning anchor), dashed $\clean$
  (clean latent). 
  Collision targets use a multi-object readout (\cref{sec:appendix:protocol}).}
  \label{fig:decodability-bars}
\end{figure}
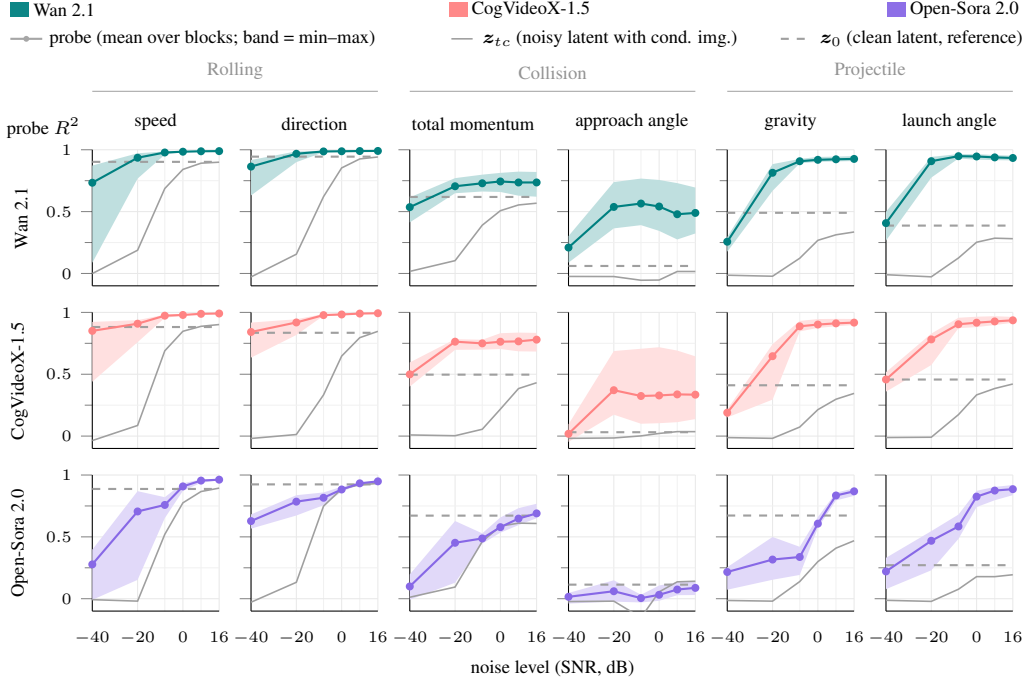

\Cref{fig:decodability-bars} reports, for every target and model, what a linear probe recovers from the \gls{dit}'s activations against what a similar probe recovers from the model's own input, along the denoising trajectory.
Every probe here reads the \gls{roi} token set, and the baseline~$\baseline$ is masked the same way, so the two differ only in what they are computed from.
Three findings hold across the grid.

At the maximum measured noise level ($-40$\,dB), the input provided to the model is uninformative, leaving the matched baseline at chance performance across all evaluated quantities. Despite this severe degradation, the model's residual stream is already capable of precisely resolving the base kinematics, while also gaining performance in the other evaluation scenarios. This indicates that \glspl{dit} actively construct physical representations over the course of the denoising trajectory, rather than passively inheriting them from the noisy input. 

As the input becomes cleaner, the physical targets split into two distinct groups, determined by the amount of information inherently carried by the clean latent. The first group consists of targets, such as rolling kinematics, where the clean latent ultimately holds the requisite information. For these properties, the model's performance margin over the baseline correspondingly shrinks as the denoising process advances. The second group comprises quantities like gravity, launch angle, and approach angle, where the clean latent carries minimal informative data. Here, the model maintains a substantial predictive margin: gravity and launch angle stay $+0.40$ to $+0.69$ above the baseline in every model even at the cleanest measured input, and approach angle $+0.47$ and $+0.30$ for Wan and CogVideoX --- though Open-Sora does not recover that target at all ($-0.05$).

Beyond the differences across physical quantities, the emergence of this information depends heavily on both the network depth and the denoising timeline. 
Across all evaluated models, decodability remains remarkably consistent throughout the network's depth: physical properties become readable within the initial transformer blocks and persist to the final output layers with peaks in early-mid layers (see \cref{sec:appendix:depth}). 
However, the models diverge significantly along the denoising trajectory. 
Wan and CogVideoX build the evaluated quantities early, reaching peak performance while the input is still heavily noised. 
In contrast, Open-Sora requires a much cleaner input, with decodability only rising near the midpoint of the diffusion process.
\Cref{tab:decodability-arms} extends this to more quantities and reports the global readout alongside the \gls{roi} readout.

\subsection{Spatio-temporal Localization}
\label{sec:probing:localization}
Having established that physical quantities can be read from the model’s activations, we next ask where this physical information is stored within the token sequence.
Unless stated otherwise, we use a fixed mid-depth layer (Layer 21) and a $0$\,dB denoising level, where all models already provide reliable physical readouts. We examine three complementary aspects: spatial localization across token positions, temporal localization across token-frames, and temporal aggregation, i.e.\ where information from several frames is combined.

\paragraph{Spatial localization}
We first ask whether physical quantities are represented at the location of the object that carries them. We compare five token sets at fixed probe, feature dimension and sample count: the single token closest to the object center (\gls{roi}), an on-object disk of $1.5\times$ the object size, two size-matched off-object decoys (one pinned at the object's reflected first-frame position, one tracking its reflected trajectory; \cref{sec:appendix:protocol}), and the global readout over the whole token grid.
We evaluate these conditions on the single-object rolling and projectile scenarios, where the off-object regions cannot overlap with another physical object.

The results in \cref{fig:loc:where} show a clear spatial concentration of the signal. The \gls{roi} consistently outperforms the global readout, despite using only roughly $1$--$2\%$ of the tokens, and it outperforms the single center token by a smaller margin ($0.04$--$0.09$ in mean $\rsq$).
Thus, the information is highly localized, but extends beyond the center token, and the radius sweep (\cref{fig:loc:disksweep}) shows this benefit saturating around the $1.5\times$ used here.

The tracked decoy, which moves like the object but on the opposite side of the grid, retains some decodability for directional quantities, so trajectory-correlated information might leak through tracking. 
The on-object region nonetheless remains substantially more informative, and pinning the decoy at its initial location removes even this trajectory information: decodability drops sharply, particularly for CogVideoX. 
The readout is therefore tied to the object's location, not to the number of pooled tokens or to its trajectory.

\paragraph{Temporal localization}
For quantities that vary within a clip, we ask whether a token-frame represents the physical state at its own point in time. 
The elastic collision gives each object exactly two speeds, before and after contact.
We assign every token-frame to its phase and fit one probe for each speed from its own phase or from the other one: the post-collision speed from pre-collision frames (\emph{anticipation}), and the pre-collision speed from post-collision frames (\emph{memory}).

As shown in \cref{fig:loc:when}, the strongest readout consistently comes from a token-frame's own phase. 
Information nevertheless extends across the collision: pre-collision token-frames retain substantial information about the post-collision speed, while post-collision token-frames retain less information about the preceding speed across all models. 
Relative to the own-frame score for the \emph{same} target, anticipation retains $80\%$, $57\%$, and $51\%$ of the information for Wan, CogVideoX, and Open-Sora, respectively, while the reverse direction retains $67\%$, $51\%$, and $10\%$. 
The two phases are nearly independent ($r = 0.22$), ruling out the possibility that anticipation merely reflects correlation. 
Thus, temporal localization is graded rather than strict: token-frames preferentially encode the physical state of their own phase, while information about other phases remains partially accessible through the model's representations.

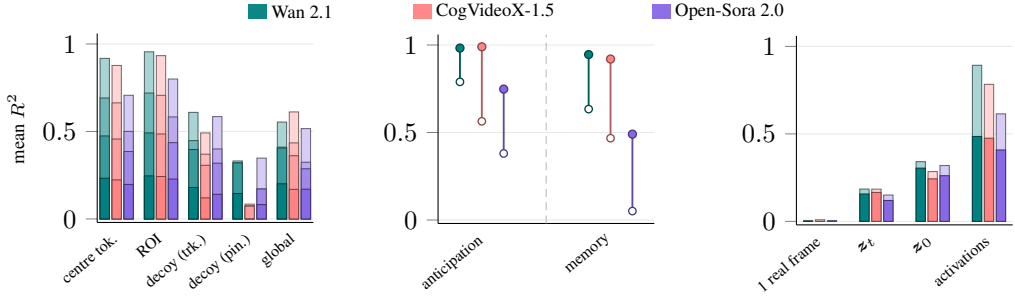
\begin{figure}[t]
  \centering
  \input{figures/probing/localization.tex}
  \caption{
    Localization of the \gls{roi} readout: \emph{spatially} \subref{fig:loc:where} and \emph{temporally} \subref{fig:loc:when} along with
    \emph{temporal aggregation} \subref{fig:loc:what}.
    Bars are probe~$\rsq$ under video-level $k$-fold cross-validation.
  }
  \label{fig:localization}
\end{figure}

\paragraph{Temporal aggregation}
In the previous experiment, we asked whether a token-frame primarily represents the physical state at its own point in time. Here, we go one step further and consider quantities for which there is no single-frame state to recover: speed and gravitational acceleration are the first and second derivatives of position and must be computed from multiple frames. While we can see in \cref{fig:decodability-bars} that these quantities are readable from the full sequence, here we want to evaluate if \glspl{dit} aggregate them across time. 

To test this, we probe four representations of a single frame: the original pixel frame, the noised latent ($\noisy$), the clean latent ($\clean$), and the corresponding \gls{dit} activation, as presented in \cref{fig:loc:what}. The pixel frame is at chance, as expected, since a single position sample contains no information about its temporal derivatives. The clean latent already contains some information because the \gls{vae} temporally compresses roughly four real frames into one latent frame --- eight for CogVideoX-1.5, whose \gls{dit} pairs latent frames ($p_t = 2$) --- making speed partially recoverable before the \gls{dit}. The \gls{dit} activations, however, recover substantially more, indicating that the transformer aggregates information beyond what is available in the local latent frame alone. This effect is particularly clear for gravity: the clean latent carries little information ($\rsq=0.07$--$0.12$), while the corresponding activations reach $\rsq=0.41$--$0.81$.
Hence, the observed temporal aggregation is not simply a consequence of the \gls{vae} compression: the \gls{dit} gathers information across frames and makes the resulting quantity locally readable.

\subsection{Generalization Across Objects and Scenes}
\label{sec:probing:generalization}
\begin{wrapfigure}{r}{0.46\textwidth}
  \centering
  \vspace{-\baselineskip}
  \input{figures/probing/generalization.tex}
  \caption{
    A \gls{roi} probe fitted on one scene and evaluated on another after recalibration.
    Hollow marks on the left edge are off-scale values below $0$.
  }
  \label{fig:probing:generalization}
  \vspace{-\baselineskip}
\end{wrapfigure}
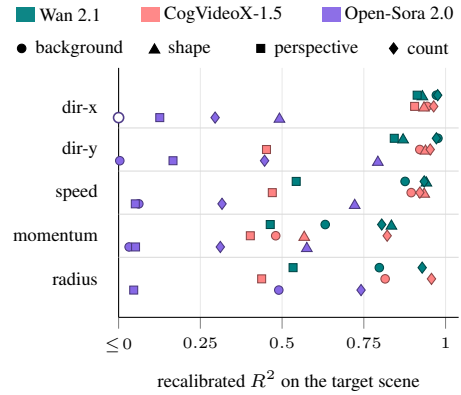
To distinguish probes that read physical quantities from those that exploit scene-specific visual regularities, we transfer probes to controlled \gls{ood} scene variations, see \cref{fig:app:scene-conditions}.
In rolling, we swap the wooden floor for a synthetic checkerboard (background), the ball for a soda can (shape), or the top-down camera for an oblique side view (perspective).
In collision, we increase the number of balls from two to five (count).

A scene change can alter a readout's scale and offset without changing its direction.
We therefore calibrate each transferred probe using only a gain~$\gain$ and offset~$\offset$ on its output, fitted by least squares on ten target-scene videos, so that it's predicted target becomes~$\pred{q} = \gain\,\probe^\top\act + \offset$.
Hence, the probe weights~$\probe$, and with them the readout direction $\probe/\lVert\probe\rVert$, remain fixed.
Please note that ten videos cannot guarantee the optimal affine calibration; hence, the appendix gives the uncalibrated scores and the calibration ceiling (\cref{sec:appendix:extrapolation}).
This approach separates transfer of the learned direction from scene-specific calibration.
For each model, we use the layer selected by pooled source-scene cross-validation and evaluate at $0$\,dB \gls{snr}.

\Cref{fig:probing:generalization} reveals a pronounced separation between the models. 
Open-Sora transfers poorly, with a median $\rsq=0.295$ compared with $0.842$ in-scene, whereas Wan and CogVideoX-1.5 retain high scores, with medians of $0.876$ and $0.905$, compared with $0.981$ and $0.977$ in-scene, respectively. 
Viewpoint changes pose the greatest challenge, while changes in object count and shape --- both without background changes --- extrapolate most reliably.
Across targets, directional quantities are generally the most robust, whereas appearance-related quantities perform worst: radius and momentum, whose mass term depends on size in our scenes.

Additional results in \cref{tab:app:generalization} show that global probes transfer through the shape change (median recalibrated $\rsq=0.773$) but collapse under every other shift (background $0.032$, perspective $0.095$, count $-0.006$). The shape swap is the only one that leaves both the background and the number of objects intact, so it is the only one under which a global mean sees nearly the same scene.
Localization therefore not only supports within-scene decodability but also enables transfer across scenes.

\input{sections/steering.tex}

%% file: figures/probing/decodability_lines_sweep.tex
\def\sweepDataFile{figures/probing/decodability_lines_sweep.dat}

\pgfmathsetlengthmacro{\sweepPanelW}{(\linewidth-51pt-60pt)/6}

\pgfplotsset{
  keep only/.style 2 args={
    x filter/.code={
      \edef\sweepKeepVal{\thisrow{#1}}
      \edef\sweepKeepMatch{#2}
      \ifx\sweepKeepVal\sweepKeepMatch\else
        \def\pgfmathresult{}
      \fi
    }
  }
}

\pgfplotsset{
  probeSweepAxis/.style={
    width=\sweepPanelW, height=1.8cm, scale only axis,
    ymin=-0.1,
    ymax=1.0,
    xmin=-40, xmax=16,
    xtick={-40,-20,0,16},
    ytick={0,0.5,1},
    minor ytick={0.25,0.75}, minor xtick={-8,8},
    yminorgrids=true, xminorgrids=true,
    minor grid style={gray!11, line width=0.3pt},
    tick label style={font=\tiny},
    label style={font=\scriptsize},
    title style={font=\scriptsize, yshift=-2pt},
    ymajorgrids=true,
    xmajorgrids=true,
    grid style={gray!18, line width=0.3pt},
    axis x line*=bottom, axis y line*=left,
    tick align=outside, xtick style={draw=none},
    unbounded coords=jump, filter discard warning=false,
    clip=true,
  },
}

\newcommand{\sweepPanel}[5]{%
  \nextgroupplot[title={#4}, #5]
    \addplot[name path=sweepLo, draw=none, forget plot]
      table[x=noise_db, y=#3_lo, keep only={quantity}{#2}] {\sweepDataFile};
    \addplot[name path=sweepHi, draw=none, forget plot]
      table[x=noise_db, y=#3_hi, keep only={quantity}{#2}] {\sweepDataFile};
    \addplot[fill=#1!25!white, draw=none, forget plot, on layer=axis background]
      fill between[of=sweepLo and sweepHi];
    \addplot[gray!75, thick, dashed, mark=none]
      table[x=noise_db, y=#3_z0, keep only={quantity}{#2}] {\sweepDataFile};
    \addplot[gray!75, semithick, mark=none]
      table[x=noise_db, y=#3_ztc, keep only={quantity}{#2}] {\sweepDataFile};
    \addplot[thick, mark=*, mark size=1.1pt, color=#1]
      table[x=noise_db, y=#3_r2, keep only={quantity}{#2}] {\sweepDataFile};}

\newcommand{\sweepPanelUnmeasured}[5]{%
  \nextgroupplot[title={#4}, #5]
    \addplot[gray!75, thick, dashed, mark=none]
      table[x=noise_db, y=#3_z0, keep only={quantity}{#2}] {\sweepDataFile};
    \addplot[gray!75, semithick, mark=none]
      table[x=noise_db, y=#3_ztc, keep only={quantity}{#2}] {\sweepDataFile};
    \node[font=\tiny, color=gray!75, align=center] at (axis cs:-12,0.72)
      {probe not\\ measured};}

\begin{tikzpicture}
  \begin{scope}[shift={(0pt,0pt)}]   \probeSwatch{modelWan}{Wan 2.1}\end{scope}
  \begin{scope}[shift={(165pt,0pt)}]  \probeSwatch{modelCog}{CogVideoX-1.5}\end{scope}
  \begin{scope}[shift={(330pt,0pt)}] \probeSwatch{modelSora}{Open-Sora 2.0}\end{scope}
  \begin{scope}[shift={(0pt,-11pt)}]
    \draw[gray!70, thick] (0,2.75pt) -- (12pt,2.75pt);
    \fill[gray!70] (6pt,2.75pt) circle (1.1pt);
    \node[anchor=west, font=\scriptsize, inner sep=1.5pt] at (13pt,2.75pt)
      {probe (mean over blocks; band = min--max)};
  \end{scope}
  \begin{scope}[shift={(196pt,-11pt)}]
    \draw[gray!75, semithick] (-30pt,2.75pt) -- (-22pt,2.75pt);
    \node[anchor=west, font=\scriptsize, inner sep=1.5pt] at (-20pt,2.75pt)
      {$\baseline$ (noisy latent with cond. img.)};
  \end{scope}
  \begin{scope}[shift={(290pt,-11pt)}]
    \draw[gray!75, thick, dashed] (0,2.75pt) -- (12pt,2.75pt);
    \node[anchor=west, font=\scriptsize, inner sep=1.5pt] at (13pt,2.75pt)
      {$\clean$ (clean latent, reference)};
  \end{scope}
\end{tikzpicture}

\newcommand{\sweepScenarioBracket}[3]{%
  \draw[gray!70, line width=0.4pt]
    ($(sweepgrp c#1r1.north west)+(0,22pt)$) --
    ($(sweepgrp c#2r1.north east)+(0,22pt)$)
    node[midway, anchor=south, font=\scriptsize, color=gray!85, yshift=1pt]
      {#3};}

\begin{tikzpicture}
  \begin{groupplot}[
    group style={group name=sweepgrp, group size=6 by 3,
                  horizontal sep=12pt, vertical sep=10pt,
                  xticklabels at=edge bottom, yticklabels at=edge left},
    probeSweepAxis,
  ]
    \sweepPanel{modelWan}{roll_speed}              {wan}{speed}                    {ylabel={Wan 2.1}}
    \sweepPanel{modelWan}{roll_direction}          {wan}{direction}                {}
    \sweepPanel{modelWan}{coll_total_momentum_xy}  {wan}{total momentum}           {}
    \sweepPanel{modelWan}{coll_approach_angle_deg} {wan}{approach angle}           {}
    \sweepPanel{modelWan}{proj_gravity}            {wan}{gravity}                  {}
    \sweepPanel{modelWan}{proj_launch_angle_deg}   {wan}{launch angle}             {}
    \sweepPanel{modelCog}{roll_speed}              {cog}{}{ylabel={CogVideoX-1.5}}
    \sweepPanel{modelCog}{roll_direction}          {cog}{}{}
    \sweepPanel{modelCog}{coll_total_momentum_xy}  {cog}{}{}
    \sweepPanel{modelCog}{coll_approach_angle_deg} {cog}{}{}
    \sweepPanel{modelCog}{proj_gravity}            {cog}{}{}
    \sweepPanel{modelCog}{proj_launch_angle_deg}   {cog}{}{}
    \sweepPanel{modelSora}{roll_speed}              {sora}{}{ylabel={Open-Sora 2.0}}
    \sweepPanel{modelSora}{roll_direction}          {sora}{}{}
    \sweepPanel{modelSora}{coll_total_momentum_xy}  {sora}{}{}
    \sweepPanel{modelSora}{coll_approach_angle_deg} {sora}{}{}
    \sweepPanel{modelSora}{proj_gravity}            {sora}{}{}
    \sweepPanel{modelSora}{proj_launch_angle_deg}   {sora}{}{}
  \end{groupplot}

  \sweepScenarioBracket{1}{2}{Rolling}
  \sweepScenarioBracket{3}{4}{Collision}
  \sweepScenarioBracket{5}{6}{Projectile}

  \node[anchor=south east, font=\scriptsize]
    at ($(sweepgrp c1r1.north west)+(-1pt,1pt)$) {probe $\rsq$};

  \node[anchor=north, font=\scriptsize] at ($(sweepgrp c1r3.south west)!0.5!(sweepgrp c6r3.south east)+(0,-15pt)$) {noise level (\glsxtrshort{snr}, dB)};
\end{tikzpicture}

%% file: figures/probing/localization.tex
\newcommand{\locBars}[2]{%
  \addplot[probeBar =modelWan,  bar shift=-\locBarStep] table[x=#1, y=r2_wan]  {#2};
  \addplot[probeBar=modelWan,  bar shift=-\locBarStep] table[x=#1, y=st_wan]  {#2};
  \addplot[probeBar =modelCog,  bar shift=0pt]          table[x=#1, y=r2_cog]  {#2};
  \addplot[probeBar=modelCog,  bar shift=0pt]          table[x=#1, y=st_cog]  {#2};
  \addplot[probeBar =modelSora, bar shift=\locBarStep]  table[x=#1, y=r2_sora] {#2};
  \addplot[probeBar=modelSora, bar shift=\locBarStep]  table[x=#1, y=st_sora] {#2};}

\newcommand{\locBarsMeasured}[2]{%
  \addplot[probeBar=modelWan,  bar shift=-\locBarStep] table[x=#1, y=r2_wan]  {#2};
  \addplot[probeBar=modelCog,  bar shift=0pt]          table[x=#1, y=r2_cog]  {#2};
  \addplot[probeBar=modelSora, bar shift=\locBarStep]  table[x=#1, y=r2_sora] {#2};}

\newcommand{\locStackModel}[5]{%
  \addplot[probeBar=#1B!34!white, bar shift=#3] table[x=#4, y=c4_#2] {#5};
  \addplot[probeBar=#1B!56!white, bar shift=#3] table[x=#4, y=c3_#2] {#5};
  \addplot[probeBar=#1B!78!white, bar shift=#3] table[x=#4, y=c2_#2] {#5};
  \addplot[probeBar=#1B,          bar shift=#3] table[x=#4, y=c1_#2] {#5};}

\newcommand{\locBarsStacked}[2]{%
  \locStackModel{modelWan}{wan}{-\locBarStep}{#1}{#2}%
  \locStackModel{modelCog}{cog}{0pt}{#1}{#2}%
  \locStackModel{modelSora}{sora}{\locBarStep}{#1}{#2}}

\newcommand{\locStackModelII}[5]{%
  \addplot[probeBar=#1B!34!white, bar shift=#3] table[x=#4, y=c2_#2] {#5};
  \addplot[probeBar=#1B,          bar shift=#3] table[x=#4, y=c1_#2] {#5};}

\newcommand{\locBarsStackedII}[2]{%
  \locStackModelII{modelWan}{wan}{-\locBarStep}{#1}{#2}%
  \locStackModelII{modelCog}{cog}{0pt}{#1}{#2}%
  \locStackModelII{modelSora}{sora}{\locBarStep}{#1}{#2}}

\newcommand{\locCeilModel}[5]{%
  \addplot[draw=#1B!55!black, fill=none, line width=0.45pt, bar shift=#3]
    table[x=#4, y=c2_#2] {#5};
  \addplot[probeBar=#1B, bar shift=#3] table[x=#4, y=c1_#2] {#5};}

\newcommand{\locBarsCeiling}[2]{%
  \locCeilModel{modelWan}{wan}{-\locBarStep}{#1}{#2}%
  \locCeilModel{modelCog}{cog}{0pt}{#1}{#2}%
  \locCeilModel{modelSora}{sora}{\locBarStep}{#1}{#2}}

\newcommand{\locDumbbell}[4]{%
  \addplot[only marks, mark=*, mark size=1.5pt, bar shift=0pt,
           draw=#1B!55!black, fill=#1B, line width=0.5pt,
           error bars/.cd, y dir=minus, y explicit, error mark=none,
           error bar style={#1B!70!black, line width=0.7pt}]
    table[x=x_#2, y=own_#2, y error=drop_#2] {#4};
  \addplot[only marks, mark=*, mark size=1.4pt, bar shift=0pt,
           draw=#1B!55!black, fill=white, line width=0.5pt]
    table[x=x_#2, y=cross_#2] {#4};}

\newcommand{\locBarsDumbbell}[1]{%
  \locDumbbell{modelWan}{wan}{arm}{#1}%
  \locDumbbell{modelCog}{cog}{arm}{#1}%
  \locDumbbell{modelSora}{sora}{arm}{#1}}

\def\barWidth{3.5pt}
\def\barGap{1pt}
\pgfmathsetlengthmacro{\locBarStep}{\barWidth+\barGap}

\pgfplotsset{
  locPanel/.style={
    probeBarAxis,
    width=0.65\textwidth, height=2.5cm,
    bar width=\barWidth,
    ymin=-0.02, ymax=1.06,
    ytick={0,0.5,1},
    label style={font=\scriptsize},
    xticklabel style={font=\tiny, rotate=40, anchor=north east,
    xshift=3pt, yshift=6pt},
    enlarge x limits=0.14,
  },
}

\begin{tikzpicture}
  \begin{scope}[shift={(0pt,0pt)}]   \probeSwatch{modelWan}{Wan 2.1}\end{scope}
  \begin{scope}[shift={(62pt,0pt)}]  \probeSwatch{modelCog}{CogVideoX-1.5}\end{scope}
  \begin{scope}[shift={(152pt,0pt)}] \probeSwatch{modelSora}{Open-Sora 2.0}\end{scope}
\end{tikzpicture}

\begin{subfigure}[t]{0.315\textwidth}
  \centering
  \begin{tikzpicture}
    \begin{axis}[locPanel,
      ylabel={mean $\rsq$},
      symbolic x coords={center, disk, rand, randstat, all},
      xtick={center, disk, rand, randstat, all},
      xticklabels={centre tok., ROI, decoy (trk.), decoy (pin.), global},
      enlarge x limits=0.11,
    ]
      \locBarsStacked{arm}{figures/probing/localization_where.dat}
    \end{axis}
  \end{tikzpicture}
  \caption{\textsc{spatial localization} --- which tokens. The \gls{roi} beats the global readout. Each bar stacks the four targets of the two single-object scenarios --- from saturated to pale, rolling speed \& direction, projectile gravity and launch angle --- at a quarter of their $\rsq$ each, so the bar height is their mean.}
  \label{fig:loc:where}
\end{subfigure}\hfill
\begin{subfigure}[t]{0.315\textwidth}
  \centering
  \begin{tikzpicture}
    \begin{axis}[locPanel,
      xmin=0.66, xmax=2.34,
      xtick={1,2},
      xticklabels={anticipation, memory},
      enlarge x limits=false,
    ]
      \locBarsDumbbell{figures/probing/localization_when.dat}
      \draw[gray!55, densely dashed, line width=0.4pt]
        (axis cs:1.5,-0.02) -- (axis cs:1.5,1.06);
    \end{axis}
  \end{tikzpicture}
  \caption{
    \textsc{temporal localization} --- which token-frame. 
    Filled dots show the target read from its own frame; hollow dots show the \emph{same} target read from the other phase of the collision. 
    Own-frame readouts are highest for every model, and anticipation consistently outperforms memory.
  }
  \label{fig:loc:when}
\end{subfigure}\hfill
\begin{subfigure}[t]{0.315\textwidth}
  \centering
  \begin{tikzpicture}
    \begin{axis}[locPanel,
      symbolic x coords={video1, zt, z0, act},
      xtick={video1, zt, z0, act},
      xticklabels={1 real frame, $\noisy$, $\clean$, activations},
    ]
      \locBarsStackedII{source}{figures/probing/localization_what.dat}
    \end{axis}
  \end{tikzpicture}
  \caption{\textsc{temporal aggregation} --- where the quantity is aggregated: the same \gls{roi} probed at a single frame, stacking rolling speed (saturated) and projectile gravity (pale) at half their $\rsq$ each. 
  Neither is readable from a single real frame and only partly from the latents but highly from the activations.}
  \label{fig:loc:what}
\end{subfigure}

%% file: figures/probing/generalization.tex
\pgfplotsset{
  genAxis/.style={
    width=0.70\linewidth, height=3.3cm, scale only axis,
    xmajorgrids=true,
    grid style={gray!18, line width=0.3pt},
    axis x line*=bottom, axis y line*=left,
    tick align=outside, ytick style={draw=none},
    tick label style={font=\scriptsize}, label style={font=\scriptsize},
    ytick={0,1,2,3,4}, y dir=reverse,
    enlarge y limits=0.14,
    clip=false,
    unbounded coords=discard,  %
  },
}

\newcommand{\genSeries}[6]{%
  \addplot[only marks, mark=#1, #2,
           draw=#3!55!black, fill=#3, line width=0.4pt,
           y filter/.code={\pgfmathparse{##1+#6}\pgfmathresult}]
    table[x=#4_#5_\genModelTag, y=idx] {figures/probing/generalization.dat};}

\newcommand{\genSeriesOut}[6]{%
  \addplot[only marks, mark=#1, #2,
           draw=#3!60!black, fill=white, line width=0.6pt,
           y filter/.code={\pgfmathparse{##1+#6}\pgfmathresult}]
    table[x=#4_#5_\genModelTag, y=idx] {figures/probing/generalization.dat};}

\newcommand{\genCondition}[4]{%
  \def\genModelTag{wan}\genSeries{#1}{#2}{modelWan}{#3}{#4}{-0.26}%
  \def\genModelTag{cog}\genSeries{#1}{#2}{modelCog}{#3}{#4}{0}%
  \def\genModelTag{sora}\genSeries{#1}{#2}{modelSora}{#3}{#4}{0.26}}

\newcommand{\genConditionOut}[4]{%
  \def\genModelTag{wan}\genSeriesOut{#1}{#2}{modelWan}{#3}{#4}{-0.26}%
  \def\genModelTag{cog}\genSeriesOut{#1}{#2}{modelCog}{#3}{#4}{0}%
  \def\genModelTag{sora}\genSeriesOut{#1}{#2}{modelSora}{#3}{#4}{0.26}}

\newcommand{\genRowRules}[2]{%
  \draw[gray!30, line width=0.3pt] (axis cs:#1,0.5) -- (axis cs:#2,0.5);
  \draw[gray!30, line width=0.3pt] (axis cs:#1,1.5) -- (axis cs:#2,1.5);
  \draw[gray!30, line width=0.3pt] (axis cs:#1,2.5) -- (axis cs:#2,2.5);
  \draw[gray!30, line width=0.3pt] (axis cs:#1,3.5) -- (axis cs:#2,3.5);}

\newcommand{\genPanel}[1]{%
  \genCondition{*}{mark size=1.6pt}{#1in}{bg}%
  \genCondition{triangle*}{mark size=2.3pt, every mark/.append style={yshift=-0.58pt}}{#1in}{shp}%
  \genCondition{square*}{mark size=1.5pt}{#1in}{psp}%
  \genCondition{diamond*}{mark size=2.1pt}{#1in}{cnt}%
  \genConditionOut{*}{mark size=1.9pt}{#1out}{bg}%
  \genConditionOut{triangle*}{mark size=2.6pt, every mark/.append style={yshift=-0.65pt}}{#1out}{shp}%
  \genConditionOut{square*}{mark size=1.8pt}{#1out}{psp}%
  \genConditionOut{diamond*}{mark size=2.4pt}{#1out}{cnt}}

\begin{tikzpicture}
  \begin{scope}[shift={(0pt,0pt)}]   \probeSwatch{modelWan}{Wan 2.1}\end{scope}
  \begin{scope}[shift={(47pt,0pt)}]  \probeSwatch{modelCog}{CogVideoX-1.5}\end{scope}
  \begin{scope}[shift={(115pt,0pt)}] \probeSwatch{modelSora}{Open-Sora 2.0}\end{scope}
  \begin{scope}[shift={(2pt,-10pt)}]
    \node[mark size=1.6pt] at (0,0) {\pgfuseplotmark{*}};
    \node[anchor=west, font=\scriptsize, inner sep=3pt] at (2pt,0) {background};
    \node[mark size=2.3pt, yshift=-0.58pt] at (50pt,0)
      {\pgfuseplotmark{triangle*}};
    \node[anchor=west, font=\scriptsize, inner sep=3pt] at (52pt,0) {shape};
    \node[mark size=1.5pt] at (90pt,0) {\pgfuseplotmark{square*}};
    \node[anchor=west, font=\scriptsize, inner sep=3pt] at (92pt,0) {perspective};
    \node[mark size=2.1pt] at (140pt,0) {\pgfuseplotmark{diamond*}};
    \node[anchor=west, font=\scriptsize, inner sep=3pt] at (142pt,0) {count};
  \end{scope}
\end{tikzpicture}

\begin{tikzpicture}
\begin{axis}[genAxis,
    xlabel={recalibrated $\rsq$ on the target scene}, xmin=0, xmax=1.04,
    xtick={0, 0.25, 0.5, 0.75, 1},
    xticklabels={$\leq\!0$, $0.25$, $0.5$, $0.75$, $1$},
    xticklabel style={font=\tiny},
    yticklabels from table={figures/probing/generalization.dat}{label},
    yticklabel style={font=\scriptsize},
]
  \genRowRules{0}{1.04}
  \genPanel{recal}
\end{axis}
\end{tikzpicture}

%% file: sections/steering.tex
\subsection{Steering}
\label{sec:steering}

Decoding a quantity from the activations shows that the representation \emph{correlates} with the physical state, not that the model uses it to produce its output.
Steering tests the latter: if pushing the activations along a probe direction reliably changes the generated motion, that direction is causal.

\paragraph{Protocol}
We steer the rolling ball's direction (left vs.\ right), the simplest target for which we obtain a usable steering direction (see \cref{sec:limitations}). At block~$\layer$, we take the steering direction $\steerDir_\layer \in \R^{\ntokens \times \dmodel}$ directly from the full-resolution probe, keeping one vector per token rather than pooling. 
We then push each token activation in $\act_{\layer}$ a fraction~$\alphastr$ toward its corresponding row of $\steerDir_\layer$ using spherical linear interpolation (SLERP).

We apply the intervention only during the high-noise phase of generation, from pure noise until the noise-level ratio reaches approximately $-20$\,dB (see \cref{app:steering}). 
To isolate the effect of steering, we always apply the rightward direction and evaluate only seeds that roll left without intervention. 
Since the seed is fixed, any rightward motion after steering is a flip caused by the intervention; we report the fraction of such flips as the \emph{direction flip ratio}.

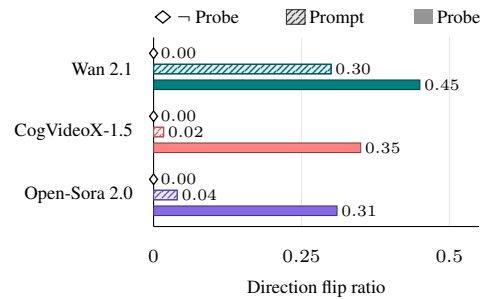
\begin{wrapfigure}[17]{r}{0.46\textwidth}
    \vspace{-\intextsep}
    \centering
    \input{figures/steering/efficacy_by_model_bars.tex}
    \vspace{-4mm}
    \caption{Direction flip ratio on left-rolling seeds, per backbone: leftward probe direction ($\neg$ Probe, negative control, marked at $0$), rightward prompt and rightward probe direction. 
    All directions are full-resolution. Configurations in \cref{tab:steering:config}.}
    \label{fig:steering:modelflipratio}
\end{wrapfigure}

\paragraph{Controls}
Two comparisons bracket the probe direction.
The \emph{leftward} probe direction is a negative control: it pushes toward the motion the seeds already have, so it should flip nothing.
A rightward \emph{prompt} --- the same seed, with the text conditioning stating that the ball rolls right --- measures how far the conditioning channel alone moves the motion.

\paragraph{Results}
The negative control behaves as intended: the leftward direction flips none of the seeds to the right on any backbone.
Prompting is at best partially effective, flipping $30\%$ of seeds for Wan but only $1.7\%$ and $4\%$ for CogVideoX and Open-Sora.
The rightward probe direction, in contrast, steers every backbone well above the control and above the prompt (\cref{fig:steering:modelflipratio}): the direction read from the activations is causally efficacious in all three models.

%% file: figures/steering/efficacy_by_model_bars.tex
\colorlet{modelWanP}{modelWanB!60!black}
\colorlet{modelCogP}{modelCogB!60!black}
\colorlet{modelSoraP}{modelSoraB!60!black}

\def\effDataFile{figures/steering/efficacy_by_model_bars.dat}

\def\effOffNeg{-0.25}
\def\effOffPrompt{0}
\def\effOffProbe{0.25}

\tikzset{effPromptHatch/.style={
  fill=#1!14!white, draw=#1!65!black,
  postaction={pattern={Lines[angle=45, distance=1.7pt, line width=0.45pt]},
    pattern color=#1!80!black}}}

\newcommand{\effBar}[6][]{%
  \addplot[probeBar=\probeColor{#2}{#3}, #1,
    nodes near coords,
    nodes near coords style={font=\tiny, anchor=west, inner sep=1.5pt,
      /pgf/number format/fixed, /pgf/number format/fixed zerofill,
      /pgf/number format/precision=2},
    point meta=x]
    table[x=#4, y expr=\coordindex+#6,
      restrict expr to domain={\coordindex}{#5:#5}]
    {\effDataFile};}
\newcommand{\effMark}[2]{%
  \addplot[only marks, mark=diamond*, mark size=2pt,
    mark options={fill=white, draw=black, line width=0.5pt},
    nodes near coords,
    nodes near coords style={font=\tiny, anchor=west, inner sep=2.5pt,
      /pgf/number format/fixed, /pgf/number format/fixed zerofill,
      /pgf/number format/precision=2},
    point meta=x]
    table[x=#1, y expr=\coordindex+\effOffNeg,
      restrict expr to domain={\coordindex}{#2:#2}]
    {\effDataFile};}

\begin{tikzpicture}
  \begin{axis}[probeBarAxis,
    xbar, bar shift=0pt,
    bar width=3.6pt,
    width=0.67\linewidth,
    height=2.5cm,
    xmin=0, xmax=0.55,
    xtick={0,0.25,0.5,0.75,1},
    xlabel={Direction flip ratio},
    xmajorgrids=true, ymajorgrids=false,
    ytick={0,1,2},
    yticklabels={Wan 2.1, CogVideoX-1.5, Open-Sora 2.0},
    y dir=reverse,
    ytick style={draw=none},
    ymin=-0.5, ymax=2.5,
    enlarge y limits=false,
    enlarge x limits=false,
    tick label style={font=\scriptsize},
    label style={font=\scriptsize},
    clip=false,
    set layers,
  ]
    \effMark{negprobe}{0}
    \effBar[effPromptHatch=modelWanB]{Wan}{P}{prompt}{0}{\effOffPrompt}
    \effBar{Wan}{B}{probe}{0}{\effOffProbe}
    \effMark{negprobe}{1}
    \effBar[effPromptHatch=modelCogB]{Cog}{P}{prompt}{1}{\effOffPrompt}
    \effBar{Cog}{B}{probe}{1}{\effOffProbe}
    \effMark{negprobe}{2}
    \effBar[effPromptHatch=modelSoraB]{Sora}{P}{prompt}{2}{\effOffPrompt}
    \effBar{Sora}{B}{probe}{2}{\effOffProbe}
  \end{axis}

  \begin{scope}[shift={($(current axis.north west)+(0pt,5pt)$)}]
    \draw[fill=white, draw=black, line width=0.5pt]
      (3.5pt,0.45pt) -- (6.3pt,2.75pt) -- (3.5pt,5.05pt) -- (0.7pt,2.75pt) -- cycle;
    \node[anchor=west, font=\scriptsize, inner sep=1.5pt] at (7pt,2.75pt)
      {$\neg$ Probe};\end{scope}
  \begin{scope}[shift={($(current axis.north west)+(50pt,5pt)$)}]
    \path[effPromptHatch=noiseRampA] (0,0) rectangle (7pt,5.5pt);
    \node[anchor=west, font=\scriptsize, inner sep=1.5pt, text depth=0pt]
      at (7pt,2.75pt) {Prompt};\end{scope}
  \begin{scope}[shift={($(current axis.north west)+(98pt,5pt)$)}]
    \probeSwatch{noiseRampB}{Probe}\end{scope}
\end{tikzpicture}

%% file: sections/discussion.tex
\section{Discussion and Limitations}
\label{sec:discussion}

\paragraph{Discussion}
The \glspl{dit} \emph{add information beyond their input}: averaged over depth, their activations outperform the noised-latent baseline.
Physical quantities are readable even at high noise, indicating that dynamics are fixed before detailed appearance.

The resulting state is \emph{computed globally but stored locally}.
Small object-centered \glspl{roi} score higher and transfer far better than the global readout, while distant tokens carry little of the relevant information.
This unsupervised object-centric representation has practical consequences.
Only a small portion of tokens suffices when the object's position is known (the simulator provides it here; real video would require tracking).
For multi-object scenes, a global readout can recover only system-level quantities such as total momentum, whereas per-object quantities require a per-object readout.
Each object also supplies a separate training sample.

The direction survives a scene change; its scale does not. 
Probes do not transfer one-to-one, because the shift changes the readout's gain and offset, but its direction is shared across the scenes we tested.
That direction is also causally active: moving activations along a direction probe changes the generated motion in every model.

Wan and CogVideoX behave similarly throughout.
Open-Sora matches them in-distribution at low noise on five of six quantities, but fails the two tests that go beyond in-distribution readout: it barely transfers, and its probe direction is the least effective for steering.
In-distribution decodability therefore does not distinguish world-model candidates; transfer against a target-scene ceiling and steering with the probe's own direction do.

Decodability alone does not establish physical understanding and benchmark failures attest to the models' limited capabilities.
The models are not trained specifically to produce physically correct video, so failure in video generation may reflect the conditioning or training mix.
Moreover, our probes read physical \emph{states}, not whether the law relating states holds.

\paragraph{Limitations}
\label{sec:limitations}
Our \emph{scope} is rigid-body mechanics in uncluttered, few-object scenes, matching the motion classes of \citet{kangHowFarVideo2025}.
We omit dissipation, rotation, deformable or articulated bodies, multi-body contact, and phenomena beyond mechanics, including fluids, thermodynamics, and optics.
A direct extension is to hidden material parameters that, like gravity, are revealed only through motion: friction, restitution, stiffness, and viscosity.

We capture activations from a \emph{forward-noised real video}, not the model's own sampling trajectory.
They are therefore off-trajectory, although capture requires no inversion.

Some targets are naturally correlated, making it hard to tell which quantity the model encodes. 
Momentum magnitude and kinetic energy correlate at $r=0.85$, and both track total mass ($r=0.71$ and $0.73$). 
All main targets are otherwise uncorrelated within each scenario.
Conversely, targets already \emph{decodable from the clean latent} cannot distinguish a model that denoises well from one that makes physics decodable.

\emph{Steering beyond direction.} Steering other physical properties remains difficult.
High-scoring probes for them are available only from pooled activations, while high-dimensional full-resolution probes fail with limited samples.
In our experiments, injecting a single vector uniformly across all tokens did not yield functional steering results, which might be due to the spatio-temporal structure of the \gls{dit} token sequence with massive activations~\citep{chengSteeringVideoDiffusion2026}.
Hence, the pooled probe family collapses when injected for steering (\cref{app:steering}).
Only steering a \gls{roi} in the token-grid, following an idea similar to that of \citet{liuTokenDialContinuousAttribute2026}, who gate tokens by a soft spatio-temporal mask, might help, but how to apply it during generation---for example, by tracking the object---remains open.

%% file: sections/conclusion.tex
\section{Conclusion}
\label{sec:conclusion}

We probed three open-weight video \glspl{dit} against simulator ground truth to ask whether they internalize physics or merely reproduce familiar motion.
\emph{Physics lives in the activations}: speed, direction, momentum and gravity are linearly decodable, the \glspl{dit} add information their input does not carry, and that information is computed globally but stored locally, at the object's own tokens.
Its readout direction survives controlled scene changes and steers the motion of generated videos --- although Open-Sora, despite matching the others in-distribution, transfers poorly.
A decodable state alone is not sufficient for physically consistent generation: whether the laws linking these states hold in the models' own representations, and why models with decodable state still fail physics benchmarks, are the questions that follow.

\subsection*{AI use statement}

In this work, we used generative AI tools to propose or refine hypotheses, design or provide feedback on research  methodology or experiments, implement methods, support qualitative and thematic data analysis, interpret results.
We have not used generative AI tools to generate synthetic data sets or to help develop theoretical models or conceptual frameworks.
Formulating mathematical claims, providing critical ingredients for proving mathematical claims, and assisting in the writing of proofs are not applicable to this work.
Additionally, we used generative AI tools to create or modify scientific figures or images, create or edit software code, draft parts of a research paper, summarize or analyze existing literature, brainstorm, identify relevant literature, format references, propose a title or keywords for a research paper. 
We have reviewed all AI-assisted work. 
All AI-assisted code was reviewed through manual pull requests and verified by our automated test pipelines before use. 
AI-suggested hypotheses, experimental designs and interpretations were critically assessed by the authors and validated against our experimental results.
All references were drawn from the authors' own curated literature library.
We take responsibility for the final content of this work, including text, claims or artifacts produced with the aid of generative AI.

\subsection*{Ethics statement}

This work does not involve human subjects, crowdsourced annotation, or personal or identifiable data. 
All videos are rendered with the Genesis physics simulator from procedurally sampled rigid-body scenes, and their labels come directly from the simulator. 
The three video models we study (Wan 2.1, CogVideoX-1.5 and Open-Sora 2.0) are publicly released open-weight checkpoints, used frozen and in accordance with their respective licenses. 
We will release our code and the rendered dataset with its simulator labels upon publication to support reproducibility; neither contains sensitive content. 
Our steering experiments show that internal activations can be edited to change generated motion. 
Activation-level control of generative video is a dual-use capability, but our interventions are limited to the lateral rolling direction of a single object in synthetic scenes and add no capability beyond existing prompt- and motion-control methods.

\subsection*{Reproducibility statement}

All three models are open-weight checkpoints, used frozen (\cref{sec:setup:models}); the simulated scenarios and their ground-truth quantities are specified in \cref{sec:setup:data,sec:simulation:scenarios}, and the probing protocol in \cref{sec:probing:method}.
Code to reproduce every experiment --- simulation, activation capture, probing, steering, and the export of every figure and table --- and the rendered videos with their simulator labels will be made available upon publication.

\ificlrfinal

\subsubsection*{Acknowledgments}
This paper has been supported by the Horizon Europe Programme (HORIZON-CL4-2022-HUMAN-02) under the project ``ELIAS: European Lighthouse of AI for Sustainability'', GA no. 101120237.
This work was funded by the National Science Centre, Poland, grant no.~UMO-2023/51/B/ST6/03004. 
We gratefully acknowledge Poland's high-performance Infrastructure PLGrid (ACC Cyfronet AGH) for providing computer facilities and support within computational grant no.~PLG/2026/019363 and PLG/2025/018644.
Created using resources provided by Wroclaw Centre for Networking and Supercomputing (\url{http://wcss.pl}).
We acknowledge the EuroHPC Joint Undertaking for awarding this project access to the EuroHPC supercomputer LEONARDO, hosted by CINECA (Italy) and the LEONARDO consortium through an EuroHPC Development Access call (project EHPC-DEV-2026D04-199).
\fi

%% file: sections/appendix.tex
\section{Data and models}

\subsection{Simulation scenarios}
\label{sec:simulation:scenarios}
All three scenarios are defined by a list of rigid objects --- each with a mass, a radius, a color, and an initial position and velocity --- as well as global scene parameters such as the gravity vector. 
Mass is tied to the rendered radius by a constant-density law, so that mass becomes visually decodable. Otherwise, absolute momentum and kinetic energy are unidentifiable from the video (non-unique) and therefore out of reach of any probe. 
The rolling scenario is stepped by the rigid-body solver, while the collision and projectile scenarios are instead rendered following the exact analytical trajectory, because Genesis' rigid contacts are effectively inelastic and hence are not suitable for our use case. 
Alongside the frames we store the per-frame position, velocity and momentum of every object, together with the scene's sampled parameters, camera pose and seeds. 
\Cref{tab:scenarios} lists what is sampled, and from where.
Each description below shows a scenario's example initial frame.

\newcommand{\scenarioThumb}[1]{%
  \includegraphics[width=\linewidth]{figures/probing/scene_#1.png}%
  }

\noindent
\begin{minipage}[t]{0.68\textwidth}
\vspace{0pt}
\textbf{Rolling without slipping}\enspace A single ball rolls across a horizontal floor without slipping. Hence, speed and heading are conserved for the whole clip and the labels are constant in time. Initial conditions are chosen in such a way that the ball does not leave the frame within the simulated time. The probe targets are the speed $\speed$ and it's direction, read out as the unit vector $\heading = \vel/\speed$.
\end{minipage}\hfill
\begin{minipage}[t]{0.3\textwidth}
\vspace{0pt}
\scenarioThumb{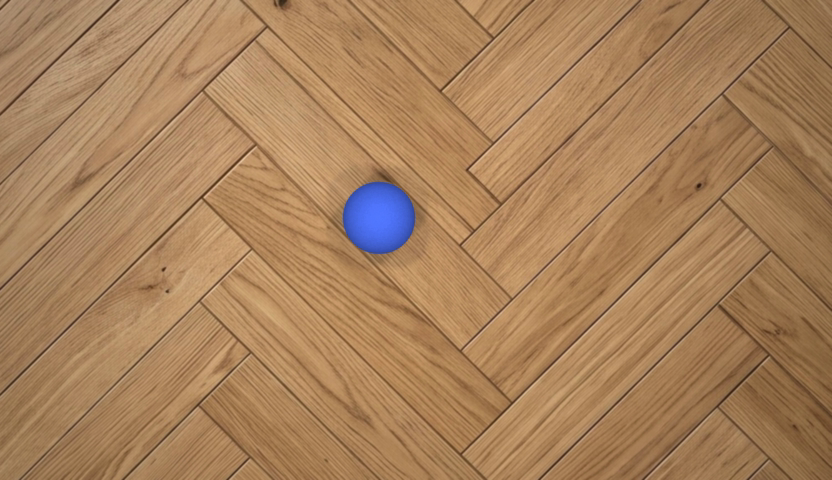}
\end{minipage}\par

\noindent
\begin{minipage}[t]{0.68\textwidth}
\vspace{0pt}
\textbf{Elastic collision}\enspace Two balls, \emph{both} in motion, meet in a frictionless elastic collision. 
They travel at constant speed between collisions. Setups that would either have a negligible impulse, and no meaningful collision, or where any of the balls leave the frame are rejected. 
The probe targets are the total momentum $\mom = \mass_1 \vel_1 + \mass_2 \vel_2$, kept as an $(x,y)$ vector, and the approach angle $\appangle$, the angle between the two velocity vectors before contact.
\end{minipage}\hfill
\begin{minipage}[t]{0.3\textwidth}
\vspace{0pt}
\scenarioThumb{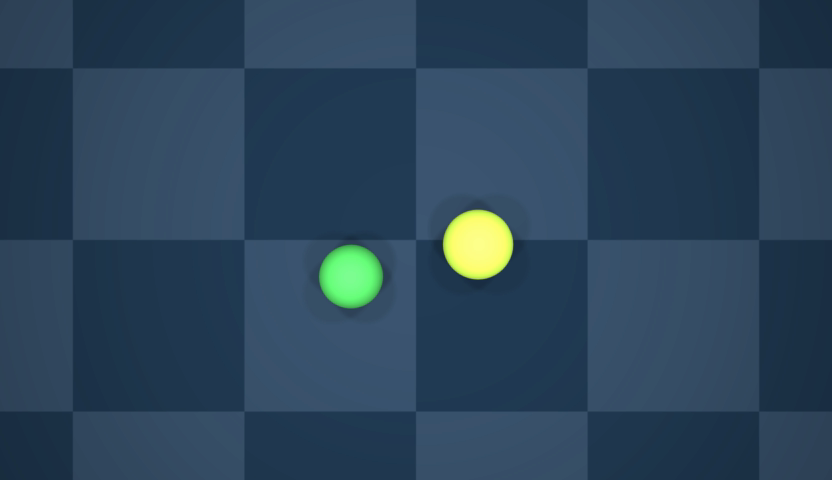}
\end{minipage}\par

\noindent
\begin{minipage}[t]{0.68\textwidth}
\vspace{0pt}
\textbf{Projectile}\enspace A single ball launches mid-air and flies a parabolic arc under a per-scene gravity magnitude $\grav$. 
The magnitude of $\grav$ is drawn from a low band so that the ball never reaches the ground within the clip. 
The launch has $x$, $y$ and $z$ components, so the path is a parabola in a vertical plane. 
The probe targets are the gravity magnitude $\grav$ and the launch angle $\launchangle$, the angle between the horizontal plane and the projectile's velocity at the moment of release.
\end{minipage}\hfill
\begin{minipage}[t]{0.3\textwidth}
\vspace{0pt}
\scenarioThumb{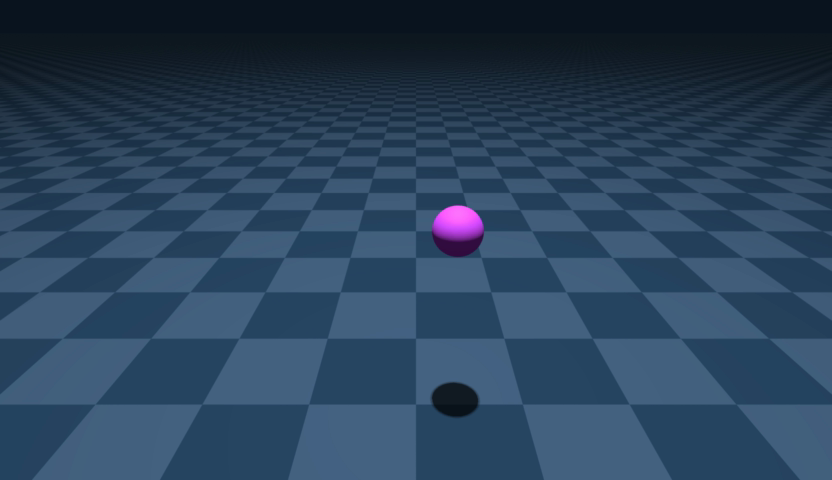}
\end{minipage}\par

\paragraph{Scene variants} Each scenario has variants that hold its sampler and physics fixed and change exactly one factor, which is what makes them usable as transfer tests (\cref{sec:probing:generalization}). 
For rolling, \emph{background} swaps the UV-mapped wood floor for an infinite checker plane; \emph{shape} replaces the ball with a cylindrical can lying on its side, rolling with the same no-slip angular velocity at a fixed radius; \emph{perspective} changes the top-down camera for an oblique side view. 
For the collision, a \emph{count} variant replaces the two-body construction with a five-object elastic hard-ball billiard on a bounded plane, invisible reflecting walls at the frame edges keeping all five in view.
The can is rendered from an analytic solution.

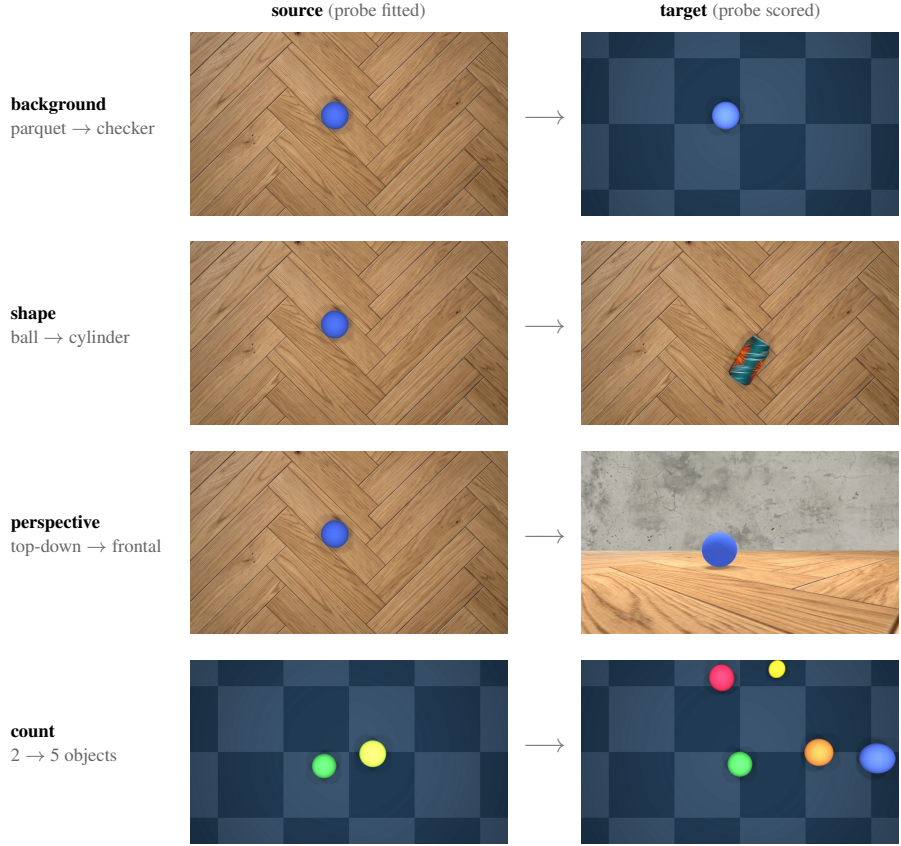
\begin{figure}[H]
  \centering
  \input{figures/probing/scene_conditions.tex}
  \caption{The four transfer shifts of \cref{sec:probing:generalization}. 
  Each row fits the probe on the left scene and scores it on the right, and changes exactly one factor between them.
  }
  \label{fig:app:scene-conditions}
\end{figure}

\begin{table}[t]
\centering
\caption{Sampled parameters per scenario. $\mathcal{U}$ denotes a uniform draw; $T$ is the clip duration and $\mass_\mathrm{ref}$, $\radius_\mathrm{ref}$ the reference mass and radius of the density law. Radius and colour are appearance nuisances in every scenario.}
\label{tab:scenarios}
\small
\begin{tabular}{llll}
\toprule
& Quantity & Symbol & Sampled from \\
\midrule
\multirow{4}{*}{\rotatebox{90}{\scriptsize Shared}}
& mass                 & $\mass$            & $\mass_\mathrm{ref}\,(\radius/\radius_\mathrm{ref})^3$, i.e.\ constant density $\rho = 1000$\,kg/m$^3$ \\
& colour               & --                  & uniform over a fixed ten-colour palette \\
& rendering            & --                  & $81$ frames at $24$\,fps ($T = 3.375$\,s) \\
& simulation step      & $\Delta t$          & $10$\,ms \\
\midrule
\multirow{4}{*}{\rotatebox{90}{\scriptsize Rolling}}
& radius               & $\radius$          & $\mathcal{U}[0.06, 0.12]$\,m \\
& speed                & $\speed$           & $\mathcal{U}[0.03, 0.12]$\,m/s \\
& heading              & $\varphi$           & $\mathcal{U}[0, 2\pi)$ \\
& start position       & $(x, y)$            & $\mathcal{U}[-0.10, 0.10]^2$\,m \\
\midrule
\multirow{7}{*}{\rotatebox{90}{\scriptsize Collision}}
& total mass           & $M = \mass_1 + \mass_2$ & $\mathcal{U}[0.3, 3.0]\,\mass_\mathrm{ref}$ \\
& mass fraction        & $\mass_1 / M$      & $\mathcal{U}[0.20, 0.80]$ \quad (radii follow from the density law) \\
& speeds               & $\norm{\vel_i}$    & $\mathcal{U}[0.04, 0.12]$\,m/s \\
& headings             & $\varphi_1$, $\varphi_2 - \varphi_1$ & $\mathcal{U}[0, 2\pi)$, $\mathcal{U}[30^\circ, 330^\circ]$ \\
& impact angle         & $\alpha$            & $\mathcal{U}[5^\circ, 75^\circ]$ away from head-on, side drawn $\pm$ \\
& contact time         & $t_c / T$           & $\mathcal{U}[0.25, 0.65]$ \\
& contact point        & $\pos_c$           & $\mathcal{U}[-0.05, 0.05]^2$\,m \\
\midrule
\multirow{7}{*}{\rotatebox{90}{\scriptsize Projectile}}
& radius               & $\radius$          & $\mathcal{U}[0.08, 0.12]$\,m \\
& gravity              & $\grav$            & $\mathcal{U}[0.14, 0.26]$\,m/s$^2$ \\
& vertical launch      & $v_{z,0}$           & $\mathcal{U}[0.24, 0.43]$\,m/s \\
& downrange launch     & $v_{x,0}$           & $\pm\,\mathcal{U}[0.08, 0.30]$\,m/s, sign drawn $50/50$ \\
& lateral launch       & $v_{y,0}$           & $\pm\,0.07$\,m/s \\
& launch angle         & $\launchangle$      & derived from $(v_{x,0}, v_{y,0}, v_{z,0})$: $39.6^\circ$--$79.4^\circ$, median $60.3^\circ$ \\
& start position       & $\pos_0$           & $\mathcal{U}\big([-0.2, 0.2] \times [-0.15, 0.15] \times [0.80, 0.90]\big)$\,m \\
\bottomrule
\end{tabular}
\end{table}

\subsection{Models and implementation details}
\label{sec:appendix:models}

\Cref{tab:models} lists the three backbones, the checkpoints we capture from,
and the clip geometry each one is run at. All three are the largest open-weight
\gls{i2v} variants of their architectures, run at their own native resolution and
frame count rather than a shared one, so that inference stays in distribution
(\cref{sec:setup:models}).

\begin{table}[H]
\centering
\caption{Backbones, checkpoints and capture geometry. \emph{Blocks} is the
number of transformer blocks the model exposes. Open-Sora's \gls{dit} is a \gls{mmdit}: its depth is 19 double-stream plus 38 single-stream blocks, indexed consecutively.}
\label{tab:models}
\footnotesize
\setlength{\tabcolsep}{3pt}
\begin{tabular}{llllll}
\toprule
Model & Checkpoint & Frames & Resolution & Blocks  \\
\midrule
Wan~2.1 (14B)        & \texttt{Wan-AI/Wan2.1-I2V-14B-480P-Diffusers} & $81$ & $480 \times 832$ & $40$ \\
CogVideoX-1.5 (5B)   & \texttt{zai-org/CogVideoX1.5-5B-I2V}          & $81$ & $768 \times 1360$ & $42$ \\
Open-Sora~2.0 (11B)  & \texttt{hpcai-tech/Open-Sora-v2}                & $81$ & $480 \times 832$ & $57$ \\
\bottomrule
\end{tabular}
\end{table}

\paragraph{Latent geometry} All three models compress by the same factor,
which is what makes their token grids comparable: each uses a causal \gls{vae}
with $8\times$ spatial and $4\times$ temporal compression that encodes the first
frame on its own, so an $81$-frame clip becomes $21$ latent frames (for Open-Sora~2.0,
in its HunyuanVAE configuration). Wan~2.1 and Open-Sora~2.0 patchify with
$p_t \timesSmall p_h \timesSmall p_w = 1 \timesSmall 2 \timesSmall 2$, giving $21$ token-frames.
CogVideoX-1.5 patchifies with $2 \timesSmall 2 \timesSmall 2$: it pads the $21$ latent
frames to $22$ by prepending a copy of the first and groups them in pairs, giving
$11$ token-frames of two latent frames each.

\paragraph{Noise levels} Captures are taken at six noise levels, reported
throughout as the \gls{snr} of the noised latent:
$\sigma \in \{0.990, 0.909, 0.715, 0.500, 0.285, 0.137\}$, i.e.\ $-40$, $-20$,
$-8$, $0$, $+8$ and $+16$\,dB. The same $\sigma$ grid is used for every model,
which is what makes the noise axis comparable across schedules.
Unless stated otherwise, localization results are read at a fixed mid-depth
block and $0$\,dB, where all three models already give reliable readouts.

\FloatBarrier
\section{Probing}

\subsection{Probing protocol details}
\label{sec:appendix:protocol}

\paragraph{Time-varying targets}
Targets that vary within the clip are probed from spatially pooled activations that keep the latent frame sequence intact, $\meanOf{\act}_{\text{sp}}\in\R^{(\nframesLat+1)\times\dmodel}$.
Latent frames then enter the regression as individual samples sharing one weight vector, so the probe stays $\dmodel$-dimensional.
Since the \gls{vae} compresses multiple video frames into a single latent frame, the physical target quantity is averaged over the corresponding time window.

\paragraph{Off-object controls}
The two decoys of \cref{sec:probing:localization} reflect the object's position through the center of the token grid, and either center a size-matched disk at the reflected position of the first frame (\textit{decoy pinned}) or follow the reflected trajectory (\textit{decoy tracked}); the tracked decoy therefore moves exactly like the object, but mirrored.
If the object starts at the center, the decoy disks are pushed outward so that they do not overlap it.
We evaluate the decoys on the single-object rolling and projectile scenarios only, where an off-object region cannot overlap another physical object.

\paragraph{Multi-object readout}
Both collision targets are read against scene-level values from all objects' \glspl{roi}, and differ in how the objects are combined: approach angle concatenates the \glspl{roi}, since a relative quantity cannot be read from a symmetric summary, while total momentum~$\mom$ averages them.

\paragraph{Input baseline}
Activations are captured in an \gls{i2v} setting from a denoising process started at a noised latent~$\noisy$ (\cref{sec:setup:models}), a compressed and noised version of the simulator's ground-truth video.
To measure the information this input already carries, we concatenate~$\noisy$ with the conditioning latent~$\condLatent\in\R^{1 \times \heightLat \times \widthLat \times \channelsLat}$, repeated across the $\nframesLat+1$ latent frames, along the channel dimension, giving the baseline signal~$\baseline\in\R^{(\nframesLat+1) \times \heightLat \times \widthLat \times 2\channelsLat}$, and train a probe using the same pooling variants --- spatial and spatio-temporal, global and \gls{roi} --- as those on the \gls{dit} activations.

\paragraph{Permutation null}
Because a probe sees far more features than videos, we verify that the protocol itself yields nothing on structureless labels: we refit every probe on targets shuffled across videos, under the same folds, penalty search and pooling, with $50$~permutations.
This leaves~$\rsq$ at chance for every quantity, model, layer and noise level, with a mean of~$-0.056$.

\subsection{Additional probing results}
\label{app:probing:additional}

\begin{table}[H]
  \centering\footnotesize
  \setlength{\tabcolsep}{4pt}
  \input{figures/probing/decodability_table.tex}
  \caption{Decodability of physical targets from \gls{roi} and global readouts, at $\sigma=0.500$ ($0$\,dB; \cref{fig:decodability-bars} gives the full noise dependence). 
  Each cell is the probe $\rsq$ averaged over 10 transformer blocks across network depth,
  block~$0$ excluded, with the gap over that arm's own $\baseline$ baseline in parentheses. 
  Rows marked~(t) are time-varying: one ridge over all (video, token-frame) pairs rather than a mean of per-frame scores. 
  Per-ball rows are averaged over the two balls, which agree to $\le 0.03$ throughout, and have no comparable global cell (\naa; multi-object). 
  }
  \label{tab:decodability-arms}
\end{table}

\subsubsection{Decodability across depth, noise and \gls{roi} size}
\label{sec:appendix:depth}
\begin{figure}[H]
  \centering
  \input{figures/probing/decodability_heatmap.tex}
  \caption{The depth $\times$ noise grid behind \cref{fig:decodability-bars}.
  Each cell is one probe ($\rsq$, $n=200$, $5$-fold video-level cross-validation) fitted at one noise level and one transformer block;
  the main figure's line is the column mean of this grid and its band the column min--max. 
  The $y$ axis is \emph{relative} depth, block index divided by $L-1$ for a network of $L$ blocks, not a raw block number: the models have $40$, $42$ and $57$ blocks, so only the fraction of
  the network is comparable. 
  Each panel samples eleven uniformly spaced blocks. 
  The small panel below each grid is that panel's matched $\baseline$ baseline on the same colour
  scale.
  }
  \label{fig:decodability-heatmap}
\end{figure}
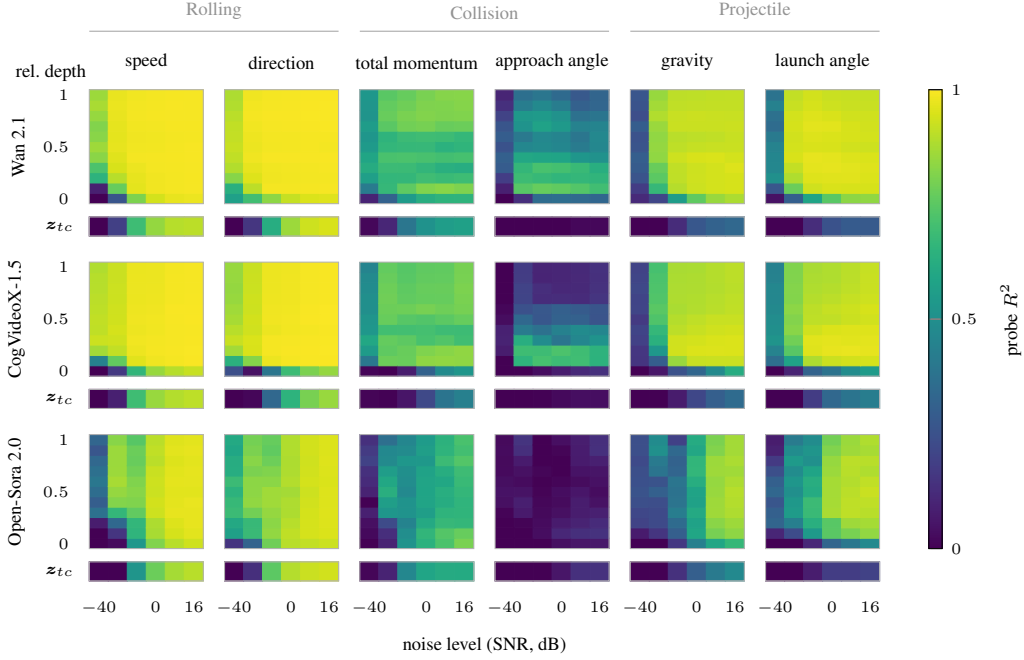

\label{sec:appendix:disksweep}
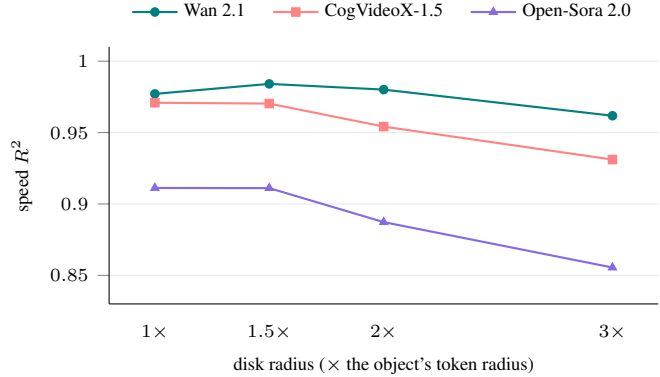
\begin{figure}[h]
  \centering
  \input{figures/probing/localization_disksweep.tex}
  \caption{Disk-radius sweep behind the single \gls{roi} bar of \cref{fig:loc:where}: speed $\rsq$ against the disk radius, as a multiple of
  the object's own token radius, at one token-frame. 
  The curve is flat between $1\times$ and
  $1.5\times$ before falling away.
  The $1.5\times$ radius is chosen as default.
  Rolling closeup-room, $n{=}200$, layer~21, $0$\,dB, all arms $\dmodel$-dimensional means over a tracked disk.}
  \label{fig:loc:disksweep}
\end{figure}

\subsubsection{Extrapolation extended results}
\label{sec:appendix:extrapolation}
Applied unchanged --- with the same weights and training-set standardization
statistics --- the probe appears not to transfer: 26 of 57 evaluations have
negative $\rsq$, and the median sits at just $0.16$ (\cref{tab:app:generalization}).
The corresponding predictions nevertheless track the targets closely, with a
median Pearson correlation of $r=0.896$. The apparent failure is therefore
mostly a calibration failure: the scene shift changes the readout's gain and
offset, while preserving much of its ordering of examples. Because $\rsq$ is
unbounded below, even a modest gain or offset mismatch can overwhelm an
otherwise informative readout.

The table separates these effects. \emph{as-is} evaluates the source probe
without adjustment; $r$ measures transfer of its ordering independent of
scale; and \emph{recal} fits only a target-scene gain and offset. The resulting
scores test whether the learned direction transfers after accounting for a
scene-specific change of units, rather than allowing the probe to be refit.

The gap between $r$ and \emph{as-is} is therefore a failure of gain and offset
rather than of direction. The two arms separate localization from decodability:
the global arm's \emph{tgt CV} is high throughout, so a global readout does
carry these quantities within a scene --- what it does not do is survive the
move to another one, with a median $r$ of $0.327$ against the \gls{roi} arm's
$0.896$. Applied unchanged, the global probe fails almost everywhere: 51 of its
57 cells are negative against 26 of 57 for the \gls{roi} arm, and the worst
reaches $-2483$.

\begin{table}[t]
  \centering\scriptsize
  \setlength{\tabcolsep}{3.5pt}
  \input{figures/probing/generalization_table.tex}
  \caption{Per-cell transfer numbers behind
  \cref{fig:probing:generalization}. 
  \emph{tgt CV} is the cross-validated score of a probe fitted \emph{and} read on the target scene: the optimal in-distribution score we could achieve. 
  \emph{as-is} applies the source probe unchanged and is unbounded below. 
  \emph{$r$} is the Pearson correlation on the target scene, which bounds what the calibration of \cref{sec:probing:generalization} could reach.
  \emph{recal} is what that ten-video calibration actually achieves --- the quantity plotted in \cref{fig:probing:generalization}, given here per cell.
  \emph{radius} is not evaluable under the object swap: the can's radius is fixed by construction, so the target has no variance --- the object-swap summary rows are therefore over four quantities, not five.}
  \label{tab:app:generalization}
\end{table}

\FloatBarrier
\section{Steering protocol details}
\label{app:steering}
\label{app:steering:protocol}

Our steering experiments push the activations of a generation along a direction associated with rightward motion and check whether the ball actually changes course. 
Below we describe the steering protocol, the text conditioning and configuration used, and what a steering direction is, how it is injected, and how its effect is scored. 
A direction can be kept at full resolution, with one vector per token, or pooled into a single vector; only the former steers (\emph{Full-resolution and pooled directions} below), and \emph{Per-backbone results} below reports what it achieves on each backbone.

\paragraph{Full-resolution and pooled directions} A \emph{full-resolution} direction at block~$\layer$ is a tensor $\steerDir_\layer\in\R^{\ntokens\times\dmodel}$ on the same token grid as the activations, with one row $\steerDir_{\layer,\tokenIndex}\in\R^{\dmodel}$ per token position $\tokenIndex=1,\dots,\ntokens$. The direction is the weight vector $\probe_\layer$ of a probe fitted on full-resolution activations, reshaped to $\ntokens\times\dmodel$; its activations come from single-step forward noising, as in the probing experiments: a labeled simulator video is encoded, noised to a chosen level and passed once through the model (\cref{sec:setup:models,sec:probing:method}). 
A \emph{pooled} direction $\meanOf{\steerDir}_\layer\in\R^{\dmodel}$ is estimated the same way from spatio-temporally mean-pooled activations, as used by the probes of \cref{sec:probing:method}, and so carries no token layout.
Pooled directions, however, do not steer. Pushes weak enough to leave the video intact keep the direction flip ratio at noise level, while stronger pushes make the video deteriorate or collapse before the ball flips. A likely reason is that the moving object occupies only a handful of the $\ntokens$ token positions: most rows $\steerDir_{\layer,\tokenIndex}$ are near zero and a few are large, so their token average, applied identically at every position, retains only a small fraction of the per-token signal ($\approx\!3\%$ of $\sum_\tokenIndex\norm{\steerDir_{\layer,\tokenIndex}}^2$ for a realized-motion contrast). Raising the push strength to compensate mostly adds a uniform drift to background tokens that carry none of the signal, which the video absorbs as a smear or global pan before the object's own motion changes. All steering results therefore use full-resolution directions.

\paragraph{Injection} A push rotates each token's activation $\act_{\layer,\tokenIndex}$ a fraction~$\alphastr\in[0,1]$ toward its target $\steerDir_{\layer,\tokenIndex}$ on the unit sphere:
\begin{equation}
    \text{\texttt{slerp}:} \quad \act_{\layer,\tokenIndex} \leftarrow \frac{\sin\big((1-\alphastr)\theta\big)}{\sin\theta}\,\act_{\layer,\tokenIndex} + \frac{\sin(\alphastr\theta)}{\sin\theta}\,\steerDir_{\layer,\tokenIndex}, \qquad \theta = \arccos\!\left(\frac{\act_{\layer,\tokenIndex}\cdot\steerDir_{\layer,\tokenIndex}}{\norm{\act_{\layer,\tokenIndex}}\,\norm{\steerDir_{\layer,\tokenIndex}}}\right) \label{eq:slerp-full}
\end{equation}
For a full-resolution direction, each token is rotated toward its own row. For a pooled direction, the target is the same vector at every position, $\steerDir_{\layer,\tokenIndex}\equiv\meanOf{\steerDir}_\layer$. The push is applied at a single block~$\layer$, during the first quarter of the denoising trajectory only (\cref{tab:steering:config}).

\paragraph{Prompts} Steered generations use a \emph{neutral} prompt that describes the scene without naming a direction; the prompt reference uses a \emph{directed} prompt that additionally states that the ball rolls right. The two differ only in the highlighted phrase:
\promptbox{Neutral prompt}{A single continuous shot: a purple sphere with a fine matte rubber-grain texture sits at rest on a warm honey-toned wood-plank floor, seen directly from above, a same wood-plank flooring extending uninterrupted to every edge of the frame, with no walls or horizon in view filling the background. The ball begins rolling smoothly sideways, accelerating horizontally until it exits the \textbf{left or right edge} of the frame.}
\promptbox{Directed prompt}{A single continuous shot: a purple sphere with a fine matte rubber-grain texture sits at rest on a warm honey-toned wood-plank floor, seen directly from above, a same wood-plank flooring extending uninterrupted to every edge of the frame, with no walls or horizon in view filling the background. The ball begins rolling smoothly sideways, accelerating horizontally until it exits the \textbf{right edge} of the frame.}

\paragraph{Evaluation} Every push is evaluated on \emph{left-rolling seeds}, whose unintervened generation produces a ball rolling left, with the direction oriented toward rightward motion. Its \emph{direction flip ratio} is the fraction of these seeds whose steered motion is classified as rightward. The classification is measured, not eyeballed. Each generated clip is passed through a Mask~R-CNN instance segmentation (COCO weights) on sampled frames; the object's mask centroid is tracked across those frames, and the resulting track is reduced to a motion class: \emph{left}, \emph{right}, or \emph{unclassified}, where the last covers ambiguous, vertical, or unreliable tracks due to degenerations or object spawns. 
A track is classified only if its longest unbroken run of detections covers at least $30\%$ of the clip and at least $70\%$ of frame pairs agree in sign with the net displacement, so a handful of large steps cannot carry a jittery track; one axis counts as dominant at a ratio of~$1.5$. Every steered clip shares its seed with an unsteered $\alphastr = 0$ clip and is compared against that clip rather than against a pooled baseline, which removes seed-to-seed variance and ensures that a seed already rolling in the target direction is never counted as a flip. Two references put the direction flip ratio in context: a \emph{prompt} reference, the flip ratio of an unsteered generation under the directed prompt (\cref{app:steering:protocol}), run with the same \gls{cfg} scale and number of denoising steps as the probe push; and a \emph{reversed} push, the same configuration rotating toward $-\steerDir_{\layer,\tokenIndex}$ instead, which should flip no seeds to the right. The configuration behind each reported result is listed in \cref{tab:steering:config}.

\paragraph{Per-backbone results}\label{app:steering:generation} \Cref{tab:steering:config} lists, for each backbone, the configuration of the probe push together with the direction flip ratio of the probe direction and of both references, the numbers \cref{fig:steering:modelflipratio} plots. All results use full-resolution directions and are evaluated on the same $52$ known left-rolling seeds for every backbone, the probe direction and both references. The probe direction exceeds the prompt reference on every backbone, and the reversed push flips none of the seeds to the right.

\begin{table}[H]
    \centering
    \caption{Steering configuration and results behind \cref{fig:steering:modelflipratio}. The first block gives the per-backbone settings of the probe push, which the negative control ($\neg$ Probe) reuses with the push reversed. The second block is shared by every generation, including the prompt reference: an unsteered generation under the directed prompt, with the same \gls{cfg} scale and number of denoising steps as the probe push. The last block gives the direction flip ratio on the $52$ left-rolling seeds, plus the fraction of seeds the reversed push leaves classified as rolling left.}
    \label{tab:steering:config}
    \small
    \begin{tabular}{l ccc}
        \toprule
        & Wan~2.1 & CogVideoX-1.5 & Open-Sora~2.0 \\
        \midrule
        Steered block $\layer$                 & $21$    & $3$    & $16$   \\
        Push strength $\alphastr$              & $0.275$ & $0.25$ & $0.25$ \\
        \midrule
        \gls{cfg} scale $\guidance$            & \multicolumn{3}{c}{$6.0$} \\
        Denoising steps                        & \multicolumn{3}{c}{$40$} \\
        Steered steps                          & \multicolumn{3}{c}{first quarter (first $10$ of $40$)} \\
        \midrule
        \multicolumn{4}{l}{\emph{Direction flip ratio}} \\
        \quad Probe                            & $0.45$  & $0.35$  & $0.31$ \\
        \quad Prompt reference                 & $0.30$  & $0.017$ & $0.04$ \\
        \quad $\neg$ Probe                     & $0.00$  & $0.00$  & $0.00$ \\
        \quad $\neg$ Probe, still classified left & $0.77$ & $0.50$ & $0.33$ \\
        \bottomrule
    \end{tabular}
\end{table}

%% file: figures/probing/scene_conditions.tex
\newcommand{\sceneCell}[1]{%
  \raisebox{-0.5\height}{%
    \includegraphics[width=0.30\linewidth]{figures/probing/scene_#1.png}}}

\newcommand{\sceneRow}[4]{%
  \parbox[c]{0.155\linewidth}{\raggedright
    \scriptsize\textbf{#1}\\[1pt]\scriptsize\textcolor{black!60}{#2}} &
  \sceneCell{#3} &
  \textcolor{black!55}{$\longrightarrow$} &
  \sceneCell{#4} \\ \addlinespace[10pt]}

\setlength{\tabcolsep}{3pt}
\begin{tabular}{@{}lccc@{}}
  & \scriptsize\textbf{source}\;\textcolor{black!60}{(probe fitted)} &
  & \scriptsize\textbf{target}\;\textcolor{black!60}{(probe scored)} \\[3pt]
  \sceneRow{background}{parquet $\to$ checker}
           {rolling-closeup-room}
           {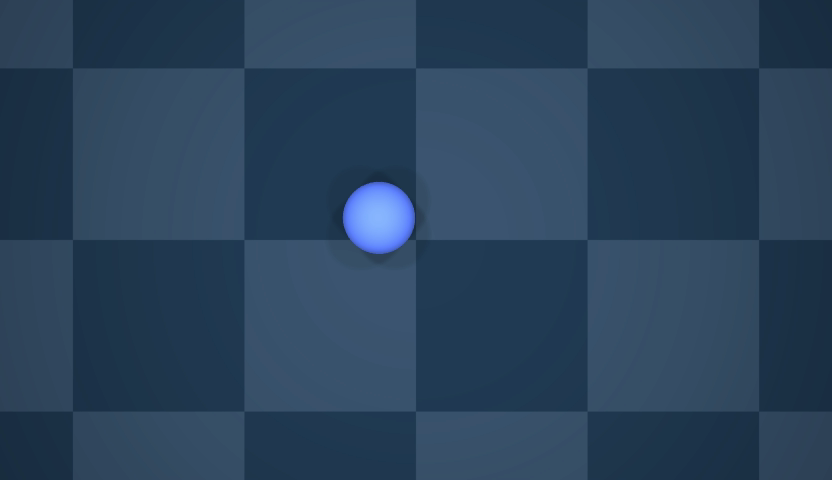}
  \sceneRow{shape}{ball $\to$ cylinder}
           {rolling-closeup-room}
           {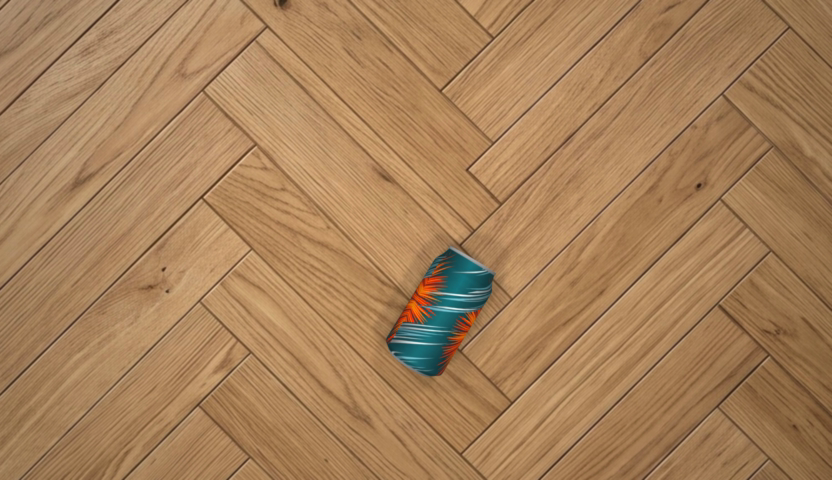}
  \sceneRow{perspective}{top-down $\to$ frontal}
           {rolling-closeup-room}
           {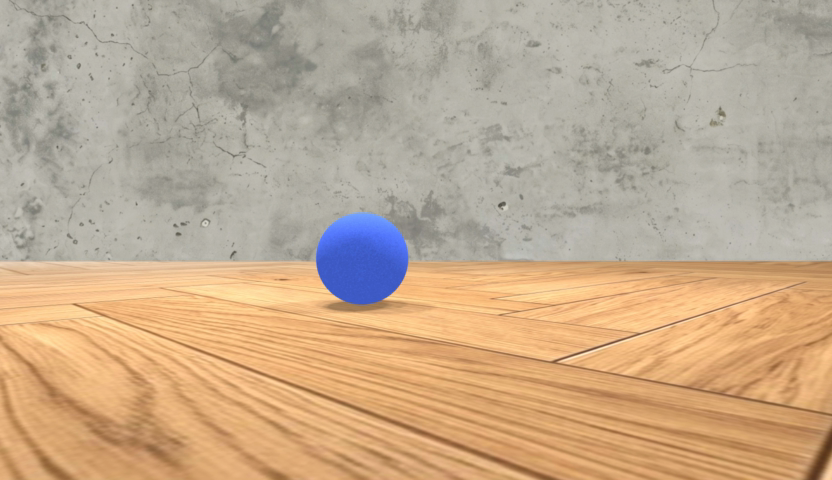}
  \sceneRow{count}{2 $\to$ 5 objects}
           {elastic-collision-continuous}
           {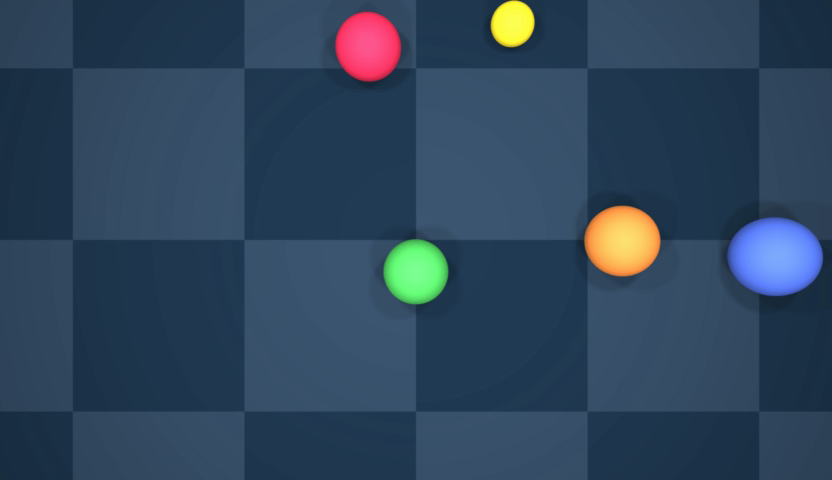}
\end{tabular}

%% file: figures/probing/decodability_table.tex
\begin{tabular}{@{}l *{6}{r}@{}}
\toprule
& \multicolumn{2}{c}{Wan 2.1} & \multicolumn{2}{c}{CogVideoX-1.5}
& \multicolumn{2}{c}{Open-Sora 2.0} \\
\cmidrule(lr){2-3} \cmidrule(lr){4-5} \cmidrule(lr){6-7}
& global & \glsxtrshort{roi} & global & \glsxtrshort{roi} & global & \glsxtrshort{roi} \\
\midrule
\multicolumn{7}{@{}l}{\textit{Rolling}} \\
  speed & 0.78 \gp{+0.53} & 0.98 \gp{+0.14} & 0.74 \gp{+0.56} & 0.98 \gp{+0.13} & 0.64 \gp{+0.49} & 0.91 \gp{+0.13} \\
  direction$_x$ & 0.85 \gp{+0.33} & 0.99 \gp{+0.07} & 0.80 \gp{+0.84} & 0.99 \gp{+0.25} & 0.53 \gp{+0.22} & 0.87 \gp{-0.01} \\
  direction$_y$ & 0.85 \gp{+0.81} & 0.99 \gp{+0.20} & 0.81 \gp{+0.90} & 0.98 \gp{+0.42} & 0.52 \gp{+0.03} & 0.90 \gp{-0.00} \\
  momentum & 0.91 \gp{+0.15} & 0.93 \gp{+0.45} & 0.92 \gp{+0.19} & 0.90 \gp{+0.40} & 0.83 \gp{+0.14} & 0.85 \gp{+0.33} \\
  kinetic energy & 0.81 \gp{+0.23} & 0.89 \gp{+0.34} & 0.85 \gp{+0.24} & 0.87 \gp{+0.32} & 0.69 \gp{+0.19} & 0.78 \gp{+0.25} \\
  mass & 0.96 \gp{+0.03} & 0.96 \gp{+0.68} & 0.93 \gp{+0.21} & 0.90 \gp{+0.54} & 0.96 \gp{+0.16} & 0.94 \gp{+0.37} \\
  radius & 0.98 \gp{+0.04} & 0.99 \gp{+0.68} & 0.94 \gp{+0.20} & 0.97 \gp{+0.56} & 0.96 \gp{+0.19} & 0.95 \gp{+0.37} \\
\addlinespace
\multicolumn{7}{@{}l}{\textit{Collision}} \\
  approach angle & 0.33 \gp{+0.34} & 0.54 \gp{+0.59} & 0.06 \gp{+0.06} & 0.33 \gp{+0.31} & -0.04 \gp{-0.02} & 0.03 \gp{-0.02} \\
  total momentum & 0.82 \gp{+0.37} & 0.74 \gp{+0.24} & 0.90 \gp{+0.83} & 0.76 \gp{+0.54} & 0.64 \gp{+0.02} & 0.58 \gp{-0.01} \\
  total kinetic energy & 0.83 \gp{+0.09} & 0.79 \gp{+0.15} & 0.93 \gp{+0.14} & 0.79 \gp{+0.17} & 0.81 \gp{+0.01} & 0.73 \gp{+0.19} \\
  total mass & 0.97 \gp{+0.06} & 0.95 \gp{+0.30} & 0.97 \gp{+0.06} & 0.91 \gp{+0.21} & 0.99 \gp{+0.07} & 0.96 \gp{+0.30} \\
  speed\,(t) & \naa & 0.95 \gp{+0.55} & \naa & 0.95 \gp{+0.70} & \naa & \textemdash \\
  direction$_x$\,(t) & \naa & 0.95 \gp{+0.42} & \naa & 0.95 \gp{+0.98} & \naa & 0.42 \gp{+0.22} \\
  direction$_y$\,(t) & \naa & 0.95 \gp{+0.96} & \naa & 0.96 \gp{+1.19} & \naa & \textemdash \\
  momentum\,(t) & \naa & 0.84 \gp{+0.56} & \naa & 0.81 \gp{+0.52} & \naa & \textemdash \\
  kinetic energy\,(t) & \naa & 0.75 \gp{+0.54} & \naa & 0.76 \gp{+0.65} & \naa & \textemdash \\
\addlinespace
\multicolumn{7}{@{}l}{\textit{Projectile}} \\
  gravity & 0.13 \gp{+0.07} & 0.92 \gp{+0.65} & 0.23 \gp{+0.30} & 0.90 \gp{+0.69} & 0.17 \gp{+0.16} & 0.61 \gp{+0.31} \\
  launch angle & 0.61 \gp{+0.53} & 0.95 \gp{+0.69} & 0.65 \gp{+0.44} & 0.92 \gp{+0.58} & 0.71 \gp{+0.62} & 0.83 \gp{+0.65} \\
  initial speed & 0.57 \gp{+0.22} & 0.90 \gp{+0.15} & 0.68 \gp{+0.34} & 0.90 \gp{+0.28} & 0.64 \gp{+0.22} & 0.80 \gp{+0.05} \\
  radius & 0.88 \gp{+0.26} & 0.89 \gp{+0.62} & 0.73 \gp{+0.33} & 0.83 \gp{+0.36} & 0.88 \gp{+0.42} & 0.89 \gp{+0.44} \\
  speed\,(t) & 0.84 \gp{+0.70} & 0.97 \gp{+0.35} & 0.83 \gp{+0.87} & 0.94 \gp{+0.65} & 0.75 \gp{+0.62} & \textemdash \\
  direction$_x$\,(t) & 0.93 \gp{+1.20} & 0.99 \gp{+0.33} & 0.93 \gp{+1.43} & 0.99 \gp{+0.89} & 0.72 \gp{+0.35} & \textemdash \\
  direction$_y$\,(t) & 0.16 \gp{+1.16} & 0.49 \gp{+1.58} & 0.14 \gp{+0.64} & 0.20 \gp{+1.10} & -0.18 \gp{+0.21} & \textemdash \\
  direction$_z$\,(t) & 0.93 \gp{+0.36} & 0.99 \gp{+0.61} & 0.94 \gp{+0.92} & 0.99 \gp{+0.58} & 0.92 \gp{+0.79} & \textemdash \\
\bottomrule
\end{tabular}

%% file: figures/probing/decodability_heatmap.tex
\def\heatDataFile{figures/probing/decodability_heatmap.dat}
\def\heatFloorFile{figures/probing/decodability_heatmap_floor.dat}

\pgfmathsetlengthmacro{\heatPanelW}{(\linewidth-46pt-40pt-54pt)/6}
\pgfmathsetlengthmacro{\heatCellW}{\heatPanelW/6}
\pgfmathsetlengthmacro{\heatCellHalf}{\heatPanelW/12}
\tikzset{heatFrame/.style={gray!60, line width=0.4pt}}

\pgfplotsset{
  probeDepthHeat/.style={
    width=\heatPanelW, height=\heatPanelW, scale only axis,
    colormap/viridis,
    point meta min=0, point meta max=1,
    axis line style={heatFrame}, axis on top,
    enlarge x limits=false, enlarge y limits=false,
    xmin=0.5, xmax=6.5, ymin=0.5, ymax=11.5,
    xtick={1,4,6}, xticklabels={},
    ytick={1,6,11}, yticklabels={$0$,$0.5$,$1$},
    tick align=outside, xtick style={draw=none}, ytick style={draw=none},
    tick label style={font=\tiny},
    label style={font=\scriptsize},
    title style={font=\scriptsize, yshift=-2pt},
  },
  probeFloorAxis/.style={
    width=\heatPanelW, height=\heatCellW, scale only axis,
    colormap/viridis, point meta min=0, point meta max=1,
    axis line style={heatFrame}, axis on top,
    enlarge x limits=false, enlarge y limits=false,
    xmin=0.5, xmax=6.5, ymin=-0.5, ymax=0.5,
    xtick={1,4,6}, xticklabels={}, ytick=\empty,
    tick align=outside, xtick style={draw=none},
    tick label style={font=\tiny}, label style={font=\scriptsize},
  },
}

\newcommand{\heatPanel}[3]{%
  \nextgroupplot[title={#2}, #3]
    \addplot[matrix plot*, point meta=explicit,
             mesh/cols=6, mesh/rows=11, mesh/ordering=y varies]
      table[x=snr_idx, y=depth, meta=#1] {\heatDataFile};}

\newcommand{\heatPanelUnmeasured}[3]{%
  \nextgroupplot[title={#2}, #3]
    \node[font=\tiny, color=gray!75, align=center] at (axis cs:3.5,6)
      {not\\ measured};}

\newcommand{\heatFloor}[4]{%
  \begin{axis}[probeFloorAxis, name=heatfl#1#2,
      at={($(heatgrp c#1r#2.south west)+(0,-5pt)$)}, anchor=north west, #4]
    \addplot[scatter, only marks, mark=square*, mark size=\heatCellHalf,
             scatter src=explicit,
             scatter/use mapped color={draw=mapped color, fill=mapped color}]
      table[x=snr_idx, y expr=0, meta=#3] {\heatFloorFile};
  \end{axis}
  \draw[heatFrame] (heatfl#1#2.south west) rectangle (heatfl#1#2.north east);}

\newcommand{\heatScenarioBracket}[3]{%
  \draw[gray!70, line width=0.4pt]
    ($(heatgrp c#1r1.north west)+(0,22pt)$) --
    ($(heatgrp c#2r1.north east)+(0,22pt)$)
    node[midway, anchor=south, font=\scriptsize, color=gray!85, yshift=1pt]
      {#3};}

\begin{tikzpicture}
  \begin{groupplot}[
    group style={group name=heatgrp, group size=6 by 3,
                 horizontal sep=8pt, vertical sep=22pt,
                 yticklabels at=edge left},
    probeDepthHeat,
  ]
    \heatPanel{wan_roll_speed}              {speed}          {ylabel={Wan 2.1}}
    \heatPanel{wan_roll_direction}          {direction}      {}
    \heatPanel{wan_coll_total_momentum_xy}  {total momentum} {}
    \heatPanel{wan_coll_approach_angle_deg} {approach angle} {}
    \heatPanel{wan_proj_gravity}            {gravity}        {}
    \heatPanel{wan_proj_launch_angle_deg}   {launch angle}   {}

    \heatPanel{cog_roll_speed}              {}{ylabel={CogVideoX-1.5}}
    \heatPanel{cog_roll_direction}          {}{}
    \heatPanel{cog_coll_total_momentum_xy}  {}{}
    \heatPanel{cog_coll_approach_angle_deg} {}{}
    \heatPanel{cog_proj_gravity}            {}{}
    \heatPanel{cog_proj_launch_angle_deg}   {}{}

    \heatPanel{sora_roll_speed}              {}{ylabel={Open-Sora 2.0}}
    \heatPanel{sora_roll_direction}          {}{}
    \heatPanel{sora_coll_total_momentum_xy}  {}{}
    \heatPanel{sora_coll_approach_angle_deg} {}{}
    \heatPanel{sora_proj_gravity}            {}{}
    \heatPanel{sora_proj_launch_angle_deg}   {}{}
  \end{groupplot}

  \heatFloor{1}{1}{wan_roll_speed}              {ylabel={$\baseline$}, ylabel style={rotate=-90, font=\tiny, xshift=2pt}}
  \heatFloor{2}{1}{wan_roll_direction}          {}
  \heatFloor{3}{1}{wan_coll_total_momentum_xy}  {}
  \heatFloor{4}{1}{wan_coll_approach_angle_deg} {}
  \heatFloor{5}{1}{wan_proj_gravity}            {}
  \heatFloor{6}{1}{wan_proj_launch_angle_deg}   {}

  \heatFloor{1}{2}{cog_roll_speed}              {ylabel={$\baseline$}, ylabel style={rotate=-90, font=\tiny, xshift=2pt}}
  \heatFloor{2}{2}{cog_roll_direction}          {}
  \heatFloor{3}{2}{cog_coll_total_momentum_xy}  {}
  \heatFloor{4}{2}{cog_coll_approach_angle_deg} {}
  \heatFloor{5}{2}{cog_proj_gravity}            {}
  \heatFloor{6}{2}{cog_proj_launch_angle_deg}   {}

  \heatFloor{1}{3}{sora_roll_speed}              {ylabel={$\baseline$}, ylabel style={rotate=-90, font=\tiny, xshift=2pt}, xticklabels={$-40$,$0$,$16$}}
  \heatFloor{2}{3}{sora_roll_direction}          {xticklabels={$-40$,$0$,$16$}}
  \heatFloor{3}{3}{sora_coll_total_momentum_xy}  {xticklabels={$-40$,$0$,$16$}}
  \heatFloor{4}{3}{sora_coll_approach_angle_deg} {xticklabels={$-40$,$0$,$16$}}
  \heatFloor{5}{3}{sora_proj_gravity}            {xticklabels={$-40$,$0$,$16$}}
  \heatFloor{6}{3}{sora_proj_launch_angle_deg}   {xticklabels={$-40$,$0$,$16$}}

  \heatScenarioBracket{1}{2}{Rolling}
  \heatScenarioBracket{3}{4}{Collision}
  \heatScenarioBracket{5}{6}{Projectile}

  \node[anchor=south east, font=\scriptsize, inner sep=1pt]
    at ($(heatgrp c1r1.north west)+(0,2pt)$) {rel.\ depth};

  \node[anchor=north, font=\scriptsize]
    at ($(heatgrp c1r3.south west)!0.5!(heatgrp c6r3.south east)+(0,-30pt)$)
    {noise level (\glsxtrshort{snr}, dB)};

  \begin{axis}[
    hide axis, scale only axis, height=0pt, width=0pt,
    colormap/viridis, point meta min=0, point meta max=1,
    colorbar,
    colorbar style={
      at={($(heatgrp c6r1.north east)+(10pt,0)$)}, anchor=north west,
      width=5pt, height=\heatPanelW*3+44pt,
      ytick={0,0.5,1}, tick label style={font=\tiny},
      ylabel={probe $\rsq$}, ylabel style={font=\scriptsize, yshift=-2pt},
    },
  ] \addplot[draw=none, forget plot] coordinates {(0,0) (1,1)};
  \end{axis}
\end{tikzpicture}

%% file: figures/probing/localization_disksweep.tex
\def\locSweepFile{figures/probing/localization_disksweep.dat}
\def\locSweepRef{figures/probing/localization_disksweep_ref.dat}

\newcommand{\locSweepModel}[3]{%
  \addplot[#1, mark=#3, mark size=1.4pt, line width=0.8pt]
    table[x=factor, y=r2_#2] {\locSweepFile};}

\begin{tikzpicture}
  \begin{axis}[probeScopeAxis,
    width=0.52\textwidth, height=3.4cm,
    xlabel={disk radius ($\times$ the object's token radius)},
    ylabel={speed $\rsq$},
    xmin=0.8, xmax=3.2,
    xtick={1.0,1.5,2.0,3.0},
    xticklabels={$1\times$,$1.5\times$,$2\times$,$3\times$},
    ymin=0.83, ymax=1.01,
    ytick={0.85,0.9,0.95,1.0},
    legend style={font=\scriptsize, draw=none, fill=none,
                  at={(0.5,1.06)}, anchor=south, legend columns=3,
                  /tikz/every even column/.append style={column sep=8pt}},
  ]
    \locSweepModel{modelWan}{wan}{*}
    \addlegendentry{Wan 2.1}
    \locSweepModel{modelCog}{cog}{square*}
    \addlegendentry{CogVideoX-1.5}
    \locSweepModel{modelSora}{sora}{triangle*}
    \addlegendentry{Open-Sora 2.0}
  \end{axis}
\end{tikzpicture}

%% file: figures/probing/generalization_table.tex
\begin{tabular}{@{}lllrrrrrrrrrrrr@{}}
\toprule
& & & \multicolumn{4}{c}{Wan 2.1} & \multicolumn{4}{c}{CogVideoX-1.5} & \multicolumn{4}{c}{Open-Sora 2.0} \\
\cmidrule(lr){4-7}\cmidrule(lr){8-11}\cmidrule(lr){12-15}
condition & quantity & arm & tgt CV & as-is & $r$ & recal. & tgt CV & as-is & $r$ & recal. & tgt CV & as-is & $r$ & recal. \\
\midrule
background & dir-x & \glsxtrshort{roi} & 0.99 & 0.95 & 0.99 & 0.97 & 0.99 & 0.89 & 0.97 & 0.94 & 0.89 & $-$14.36 & 0.32 & $-$0.03 \\
 &  & global & 0.91 & $-$63.82 & 0.53 & 0.24 & 0.95 & $-$33.11 & 0.33 & 0.08 & 0.84 & $-$1110 & 0.06 & $-$0.03 \\
\addlinespace[2pt]
 & dir-y & \glsxtrshort{roi} & 0.99 & 0.95 & 0.99 & 0.98 & 0.99 & 0.79 & 0.96 & 0.92 & 0.88 & $-$13.14 & 0.31 & 0.00 \\
 &  & global & 0.91 & $-$4.83 & 0.56 & 0.27 & 0.95 & $-$149 & 0.66 & 0.37 & 0.80 & $-$155 & 0.04 & $-$0.04 \\
\addlinespace[2pt]
 & speed & \glsxtrshort{roi} & 0.99 & 0.42 & 0.94 & 0.88 & 0.99 & 0.55 & 0.95 & 0.89 & 0.93 & $-$30.70 & 0.42 & 0.06 \\
 &  & global & 0.84 & $-$664 & 0.23 & 0.04 & 0.90 & $-$229 & 0.24 & 0.02 & 0.85 & $-$4.14 & 0.27 & $-$0.28 \\
\addlinespace[2pt]
 & momentum & \glsxtrshort{roi} & 0.93 & 0.58 & 0.80 & 0.63 & 0.89 & 0.11 & 0.71 & 0.48 & 0.93 & $-$185 & 0.34 & 0.03 \\
 &  & global & 0.91 & $-$288 & 0.18 & 0.03 & 0.93 & $-$1363 & 0.12 & 0.01 & 0.93 & $-$1104 & 0.19 & 0.04 \\
\addlinespace[2pt]
 & radius & \glsxtrshort{roi} & 0.99 & $-$0.98 & 0.90 & 0.80 & 0.98 & 0.47 & 0.91 & 0.82 & 1.00 & $-$103 & 0.71 & 0.49 \\
 &  & global & 0.99 & $-$848 & 0.51 & 0.21 & 0.97 & $-$2483 & 0.16 & 0.02 & 1.00 & $-$12.81 & 0.24 & 0.03 \\
\addlinespace[2pt]
 & \textit{median} & \textit{\glsxtrshort{roi}} & 0.99 & 0.58 & 0.94 & 0.88 & 0.99 & 0.55 & 0.95 & 0.89 & 0.93 & $-$30.70 & 0.34 & 0.03 \\
 &  & \textit{global} & 0.91 & $-$288 & 0.51 & 0.21 & 0.95 & $-$229 & 0.24 & 0.02 & 0.85 & $-$155 & 0.19 & $-$0.03 \\
\midrule
shape & dir-x & \glsxtrshort{roi} & 1.00 & 0.80 & 0.97 & 0.93 & 1.00 & 0.51 & 0.97 & 0.93 & 0.98 & $-$3.02 & 0.75 & 0.49 \\
 &  & global & 0.97 & 0.54 & 0.95 & 0.85 & 0.98 & 0.53 & 0.97 & 0.93 & 0.97 & $-$0.00 & 0.74 & 0.37 \\
\addlinespace[2pt]
 & dir-y & \glsxtrshort{roi} & 1.00 & 0.82 & 0.95 & 0.87 & 1.00 & 0.77 & 0.97 & 0.94 & 0.99 & $-$25.25 & 0.90 & 0.79 \\
 &  & global & 0.97 & 0.17 & 0.96 & 0.93 & 0.99 & 0.81 & 0.98 & 0.95 & 0.99 & $-$32.09 & 0.81 & 0.62 \\
\addlinespace[2pt]
 & speed & \glsxtrshort{roi} & 0.99 & $-$2.27 & 0.97 & 0.94 & 0.99 & $-$3.29 & 0.97 & 0.93 & 0.98 & $-$25.84 & 0.85 & 0.72 \\
 &  & global & 0.92 & $-$0.66 & 0.89 & 0.76 & 0.97 & $-$3.07 & 0.92 & 0.82 & 0.98 & $-$97.57 & 0.62 & 0.33 \\
\addlinespace[2pt]
 & momentum & \glsxtrshort{roi} & 0.99 & 0.75 & 0.92 & 0.83 & 0.99 & 0.28 & 0.85 & 0.57 & 0.98 & $-$1.12 & 0.77 & 0.58 \\
 &  & global & 0.92 & 0.37 & 0.76 & 0.34 & 0.97 & 0.68 & 0.92 & 0.79 & 0.98 & $-$1.96 & 0.57 & 0.32 \\
\addlinespace[2pt]
 & radius & \glsxtrshort{roi} & -- & -- & -- & -- & -- & -- & -- & -- & -- & -- & -- & -- \\
 &  & global & -- & -- & -- & -- & -- & -- & -- & -- & -- & -- & -- & -- \\
\addlinespace[2pt]
 & \textit{median} & \textit{\glsxtrshort{roi}} & 1.00 & 0.77 & 0.96 & 0.90 & 1.00 & 0.40 & 0.97 & 0.93 & 0.98 & $-$14.13 & 0.81 & 0.65 \\
 &  & \textit{global} & 0.94 & 0.27 & 0.92 & 0.80 & 0.98 & 0.61 & 0.94 & 0.88 & 0.98 & $-$17.02 & 0.68 & 0.35 \\
\midrule
perspective & dir-x & \glsxtrshort{roi} & 0.97 & 0.92 & 0.97 & 0.91 & 0.97 & 0.87 & 0.96 & 0.91 & 0.83 & $-$33.13 & 0.49 & 0.13 \\
 &  & global & 0.87 & $-$2.47 & 0.76 & 0.46 & 0.89 & $-$6.83 & 0.63 & 0.26 & 0.59 & $-$146 & 0.02 & $-$0.13 \\
\addlinespace[2pt]
 & dir-y & \glsxtrshort{roi} & 0.96 & 0.55 & 0.92 & 0.84 & 0.95 & 0.34 & 0.71 & 0.45 & 0.78 & $-$2.29 & 0.48 & 0.17 \\
 &  & global & 0.70 & $-$11.76 & 0.38 & 0.05 & 0.80 & $-$7.26 & 0.15 & $-$0.10 & 0.57 & $-$146 & 0.04 & $-$0.20 \\
\addlinespace[2pt]
 & speed & \glsxtrshort{roi} & 0.91 & 0.05 & 0.75 & 0.54 & 0.92 & $-$0.38 & 0.69 & 0.47 & 0.55 & $-$21.79 & 0.27 & 0.05 \\
 &  & global & 0.59 & $-$321 & 0.58 & 0.30 & 0.79 & $-$97.57 & 0.38 & 0.11 & 0.20 & $-$984 & 0.01 & $-$0.03 \\
\addlinespace[2pt]
 & momentum & \glsxtrshort{roi} & 0.78 & $-$34.39 & 0.69 & 0.46 & 0.79 & $-$19.48 & 0.65 & 0.40 & 0.66 & $-$140 & 0.31 & 0.05 \\
 &  & global & 0.73 & $-$7.13 & 0.62 & 0.37 & 0.70 & $-$20.56 & 0.52 & 0.25 & 0.42 & $-$748 & 0.22 & $-$0.10 \\
\addlinespace[2pt]
 & radius & \glsxtrshort{roi} & 0.93 & $-$24.90 & 0.74 & 0.53 & 0.91 & $-$15.06 & 0.68 & 0.44 & 0.96 & $-$29.19 & 0.37 & 0.05 \\
 &  & global & 0.92 & $-$290 & 0.50 & 0.24 & 0.82 & $-$629 & 0.33 & 0.09 & 0.93 & $-$384 & 0.22 & $-$0.25 \\
\addlinespace[2pt]
 & \textit{median} & \textit{\glsxtrshort{roi}} & 0.93 & 0.05 & 0.75 & 0.54 & 0.92 & $-$0.38 & 0.69 & 0.45 & 0.78 & $-$29.19 & 0.37 & 0.05 \\
 &  & \textit{global} & 0.73 & $-$11.76 & 0.58 & 0.30 & 0.80 & $-$20.56 & 0.38 & 0.11 & 0.57 & $-$384 & 0.04 & $-$0.13 \\
\midrule
count & dir-x & \glsxtrshort{roi} & 0.98 & 0.98 & 0.99 & 0.98 & 0.98 & 0.96 & 0.98 & 0.96 & 0.62 & $-$0.04 & 0.58 & 0.30 \\
 &  & global & $-$0.18 & $-$4.67 & 0.32 & $-$0.01 & $-$0.36 & $-$1.34 & 0.26 & 0.00 & $-$0.65 & $-$3.58 & 0.08 & $-$0.07 \\
\addlinespace[2pt]
 & dir-y & \glsxtrshort{roi} & 0.98 & 0.97 & 0.99 & 0.97 & 0.98 & 0.95 & 0.98 & 0.95 & 0.70 & 0.16 & 0.67 & 0.45 \\
 &  & global & $-$0.17 & $-$0.91 & 0.40 & $-$0.19 & $-$0.34 & $-$4.43 & 0.23 & $-$0.53 & $-$0.14 & $-$2.45 & 0.24 & $-$0.42 \\
\addlinespace[2pt]
 & speed & \glsxtrshort{roi} & 0.96 & 0.92 & 0.97 & 0.93 & 0.96 & 0.92 & 0.96 & 0.92 & 0.60 & $-$2.07 & 0.57 & 0.32 \\
 &  & global & $-$0.62 & $-$0.72 & 0.13 & $-$0.08 & $-$0.55 & $-$0.83 & 0.14 & 0.00 & $-$0.38 & $-$5.03 & 0.13 & $-$0.14 \\
\addlinespace[2pt]
 & momentum & \glsxtrshort{roi} & 0.91 & 0.36 & 0.91 & 0.80 & 0.90 & 0.69 & 0.91 & 0.82 & 0.71 & $-$6.18 & 0.60 & 0.31 \\
 &  & global & 0.29 & $-$14.48 & 0.36 & 0.04 & 0.25 & $-$9.14 & 0.29 & $-$0.01 & 0.46 & $-$10.26 & 0.31 & 0.03 \\
\addlinespace[2pt]
 & radius & \glsxtrshort{roi} & 0.98 & 0.73 & 0.97 & 0.93 & 0.98 & 0.90 & 0.98 & 0.96 & 0.93 & $-$0.57 & 0.89 & 0.74 \\
 &  & global & 0.90 & $-$7.59 & 0.37 & 0.04 & 0.69 & $-$12.16 & 0.28 & 0.01 & 0.96 & $-$5.05 & 0.12 & $-$0.02 \\
\addlinespace[2pt]
 & \textit{median} & \textit{\glsxtrshort{roi}} & 0.98 & 0.92 & 0.97 & 0.93 & 0.98 & 0.92 & 0.98 & 0.95 & 0.70 & $-$0.57 & 0.60 & 0.32 \\
 &  & \textit{global} & $-$0.17 & $-$4.67 & 0.36 & $-$0.01 & $-$0.34 & $-$4.43 & 0.26 & 0.00 & $-$0.14 & $-$5.03 & 0.13 & $-$0.07 \\
\midrule
\addlinespace[2pt]
 & \textit{all shifts} & \textit{\glsxtrshort{roi}} & 0.98 & 0.73 & 0.95 & 0.88 & 0.98 & 0.55 & 0.96 & 0.91 & 0.89 & $-$14.36 & 0.57 & 0.30 \\
 &  & \textit{global} & 0.90 & $-$7.13 & 0.51 & 0.24 & 0.89 & $-$9.14 & 0.33 & 0.08 & 0.84 & $-$32.09 & 0.22 & $-$0.03 \\
\bottomrule
\end{tabular}